\documentclass[final,12pt]{elsarticle}

\usepackage{tabularx}
\usepackage{bm}
\usepackage{textcomp}
\usepackage{xcolor}
\usepackage{float}
\usepackage{multicol}
\usepackage{multirow}
\usepackage{svg}
\usepackage[normalem]{ulem}
\usepackage[inline]{enumitem}
\usepackage{subcaption}
\restylefloat{table}
\usepackage{booktabs}
\usepackage{array}
\usepackage[linesnumbered,ruled,vlined]{algorithm2e}

\newcolumntype{P}[1]{>{\centering\arraybackslash}p{#1}}

\SetKwInput{KwInput}{Input}                
\SetKwInput{KwOutput}{Output}              
\usepackage{amsfonts}
\usepackage{amssymb} 
\usepackage{amsmath}
\usepackage{amsthm}
\usepackage{latexsym}
\usepackage{graphicx}
\usepackage{algpseudocode}
\usepackage[hidelinks]{hyperref}

\journal{Internet of Things}
\begin{document}

\begin{frontmatter}

\title{
FLAGS Framework for Comparative Analysis of Federated Learning Algorithms
}

\author{Ahnaf Hannan Lodhi\corref{cor1}}
\ead{alodhi18@ku.edu.tr}
\cortext[]{This work was supported by the Ko\c{c} University and I\c{s} Bank Artificial Intelligence (KUIS AI) Center Research Award and in part by the TUBITAK 2247-A Award 121C338. Funding Agency: TUBITAK 2247-A Award (Project No: 121C338).}
\cortext[cor1]{Corresponding author at: Graduate School of Science and Engineering, Ko\c{c} University, Rumeli Feneri, Sar{\i}yer, Istanbul 34450, Turkey}
\author{Barı\c{s} Akg\"{u}n}
\ead{baakgun@ku.edu.tr}
\author{\"{O}znur \"{O}zkasap}
\ead{oozkasap@ku.edu.tr}
\address{Department of Computer Engineering. Ko\c{c} University, Rumelifeneri} \address{Sar{\i}yer, 34450, Turkey}

\begin{abstract}
Federated Learning (FL) has become a key choice for distributed machine learning. Initially focused on centralized aggregation, recent works in FL have emphasized greater decentralization to adapt to the highly heterogeneous network edge. 
Among these, Hierarchical, Device-to-Device and Gossip Federated Learning (HFL, D2DFL \&  GFL respectively) can be considered as foundational FL algorithms employing fundamental aggregation strategies. A number of FL algorithms were subsequently proposed employing multiple fundamental aggregation schemes jointly. Existing research, however, subjects the FL algorithms to varied conditions and gauges the performance of these algorithms mainly against Federated Averaging (FedAvg) only. This work consolidates the FL landscape and offers an objective analysis of the major FL algorithms through a comprehensive cross-evaluation for a wide range of operating conditions. In addition to the three foundational FL algorithms, this work also analyzes six derived algorithms. To enable a uniform assessment, a multi-FL framework named FLAGS: \underline{F}ederated \underline{L}earning \underline{A}l\underline{G}orithms \underline{S}imulation has been developed for rapid configuration of multiple FL algorithms. 
Our experiments indicate that fully decentralized FL algorithms achieve comparable accuracy under multiple operating conditions, including asynchronous aggregation and the presence of stragglers. Furthermore, decentralized FL can also operate in noisy environments and with a comparably higher local update rate. However, the impact of extremely skewed data distributions on decentralized FL is much more adverse than on centralized variants. The results indicate that it may not be necessary to restrict the devices to a single FL algorithm; rather, multi-FL nodes may operate with greater efficiency. 
\end{abstract}

\begin{keyword}
Federated Learning \sep Hierarchical \sep Device-to-Device (D2D) \sep Gossip \sep Cluster \sep Framework
\end{keyword}
\end{frontmatter}

\section{Introduction}

The critical challenges associated with centralization in Machine Learning (ML), including data aggregation, privacy and security risks, and economy of resources have accelerated interest in decentralized learning approaches. 
Federated Learning (FL) is one such learning framework that enables disjoint devices to learn collaboratively without the need to share data. The process entails devices to train models locally and share them for aggregation. Since its inception by Google researchers, Federated Learning (FL) proposed in \cite{FL_McMahan_Moore_Ramage_Hampson_Arcas_2017} as FedAvg has been the foremost avenue for distributed Deep Learning (DL). Gboard-Google keyboard and Siri-Apple smart assistant, are real-world applications that have benefited tremendously from FL \cite{kairouz2019advances}.

While FL is a major avenue towards democratized learning, there still exists a strong proclivity in FL research towards centralized orchestration. Access to updates from a large number of devices, a global model, considerable computational and storage resources have traditionally made a strong case for Centralized FL (FedAvg). However, there exist a number of challenges \cite{kairouz2019advances} to such an operation  including adversarial attacks, malicious participation, privacy preservation, heterogeneous devices and participation, non-Independently and Identically Distributed (non-IID) data, and communication issues. Simultaneously, next generation networks \cite{ansari20175g} envision considerable serverless interaction between the devices. Advances, such as application specific data rates, Network Function Virtualization (NFV) and network slicing have enabled researchers to propose various FL algorithms based on the permissible device interactions
(see Fig.\ref{fig:networklayers} for the depiction of these links). The proposed methods under various settings show multiple advantages over Centralized FL (FedAvg). Furthermore, current and future devices come equipped with multiple communication links. This fact negates the limitation of operating a single FL algorithm at all times. The challenge then arises regarding the selection of FL algorithm to employ under various conditions. The comparative characterization of FL algorithms conducted in this work is aimed at bridging this gap. 
The analysis conducted here paves the way 
for establishing economy of operation for these algorithms for various operating scenarios in a particularly heterogeneous Network Edge.

\paragraph*{Contributions} This work aims to contrast the performance of the major FL algorithms 
under some of the most prevalent challenges to distributed learning at the network edge, i.e., non-IID data, noisy communication and asynchronous aggregation. The main goal is to  pave the way for a multi-FL system where devices may select the most suitable FL algorithm depending on the operating characteristics. 
\begin{enumerate}
\item A detailed comparative evaluation of the fundamental Federated Learning algorithms namely 
\begin{enumerate*}[label = (\alph*)]
    \item \textit{Hierarchical FL (HFL)}
    \item \textit{Device-to-Device FL (D2DFL)}
    \item \textit{Gossip FL (GFL)}
    \item \textit{Centralized FL (FedAvg)} 
    \end{enumerate*}
has been conducted. In order to expand the scope of evaluation, this work also considers other FL algorithms employing multiple aggregation strategies jointly. These include:
\begin{enumerate*}[label = (\alph*)]
    \item \textit{Hierarchical Device-to-Device FL (HD2DFL)}
    \item \textit{Hierarchical Gossip FL (HGFL)}
    \item \textit{Clustered Device-to-Device FL (CD2DFL)}
    \item \textit{Inter-Cluster FL (iCFL)}
    \item \textit{Inter-Cluster Device-to-Device FL (iCD2DFL)}
\end{enumerate*}

Their performance is tested under ideal and noisy Device-Device (D2D), Device-Edge (D2E) and Device/Edge-Cloud (D2C/E2C) communication links. 
The results indicate that even in the presence of non-IID data, 
decentralized FL performs with a marginal loss in performance compared to centralized variants despite operating entirely in the absence of any global information. Extremely skewed data distributions, however, greatly impact decentralized FL drastically reducing convergence rate as well as accuracy. 

\item A detailed study of these algorithms under various levels of device participation (such as active, inactive, stragglers) has been conducted. Decentralized FL shows marginal loss in performance when compared with centralized FL in the presence of degraded participation. Decrease in device participation and extremely skewed data distributions have a confounding affect on all the algorithms, more so on the decentralized ones. 

\item An analysis using modified Few-Shot Learning, with considerably more local updates before aggregation, has also been conducted for the FL algorithms under consideration. Decentralized algorithms with show a greater loss in accuracy than the centralized ones. However, clustered operation helps mitigate the absence of a global server while restricting all communications to the device level.
\item Finally, this work develops the FLAGS framework\footnote{\url{https://github.com/ahnaflodhi/FLAGS-FL}} to simulate multiple FL algorithms. The framework is designed to support ease of configuration of multiple FL algorithms and provides numerous options for generating the edge network topology for a realistic assessment. 
\end{enumerate}

The remainder of this paper is organized as follows: Section-\ref{sec:relatedwork} provides some of the related developments in FL. An overview of the system model and FL is presented in Section-\ref{sec:sysmodel}. The FLAGS framework and its modules have been explained in Section-\ref{sec:flags}. Details of the FL algorithms have been provided in Section-\ref{sec:FLclasses}. This is followed by a description of the experiments and performance analysis in Section-\ref{sec:expeval} and \ref{sec:analysis}. The paper concludes with a summary of the important findings in Section-\ref{sec:conclusion}. 
\section{Related Work} \label{sec:relatedwork}

The FL process is envisioned for a highly heterogeneous environment. The search of optimal application of FL has resulted in various FL algorithms designed to uniquely benefit from edge network topology. The impact of some of the most prevalent among these must first be characterized in order to adapt FL accordingly. To this end, some of the representative works that have focused on the FL algorithms and simulation frameworks have been covered in this section. 

\subsection{Federated Learning Algorithms}
Most of the current research in Federated Learning leans heavily towards centralization. However, research indicates that careful selection of design parameters may yield competitive results for decentralized FL as well. \cite{fullyFL_Lalitha_Javidi_Shekhar_Koushanfar} describes a fully decentralized FL algorithm with {IID} data distributed across users interacting according to a directed graph. In \cite{gossip_FL_2019}, the authors present a gossip-aggregation framework for FL. The research reports more favorable results from gossip learning based FL with uniform data distribution across the nodes. More recently, \cite{D2D_Wireless} suggests the application of the Decentralized Stochastic Gradient Descent (\textit{DSGD}) \cite{ram2010d2d} for D2D belief aggregation in  the presence of wireless impairments. \cite{Savazzi_Nicoli_Rampa_Kianoush_2020} proposes a Consensus-based Federated Averaging (CFA) for dense IoT networks. Their analysis suggests that serverless cooperation between devices may also yields results comparable to FedAvg. Extending their work, the authors in \cite{Savazzi_Nicoli_Rampa_2020} evaluate Consensus-based Federated Averaging (CFA) and CFA-Gradient Exchange (CFA-GE) for dense IoT networks with D2D interaction. Their results support the original hypothesis indicating that although slow to converge, CFA and CFA-GE achieve performance similar to FedAvg with communication restricted only to device level. In \cite{Li_Li_Varshney_2022}, the authors try to tackle the problem of data heterogeneity in a decentralized fit learning setting. The authors propose a peer-to-peer model exchange method with model fusion using Mutual Knowledge Transfer.

Other research has also proposed hybrid FL algorithms employing hierarchical aggregation in conjunction with D2D interaction. In \cite{Wang_Wang_Chen_Ji_2021}, the authors perform divergence based client grouping in a Hierarchical Federated Learning (HFL) scenario. \cite{Abad_Ozfatura_GUndUz_Ercetin_2020} provide latency analysis of a hierarchical FL system operating in a heterogeneous cellular network where the local aggregation is conducted at Mobile Base Stations. The impact of HFL on training time and energy consumption is investigated in \cite{liu2020client}. The same work also indicates that a trade-off between latency and computation may be achieved resulting in better performance than centralized FL. A two-time scale Hybrid FL model is proposed in \cite{Lin_Hosseinalipour_Azam_Brinton_Michelusi_2021} which compliments device-device communication with hierarchical server based aggregations. This work introduces a control algorithm scheduling global aggregations, local interactions and learning rate to achieve a convergence rate of $\mathcal{O}(1/t)$.
The work in \cite{hashemi2021benefits} uses gossip interaction between the devices before allowing them to upload their models to the hierarchical servers. The resultant FL algorithm provides near-optimal results even in the presence of reduced communication frequency and volume. \cite{wang2021edge} proposes clustering devices in the same network location using the corresponding mobile edge nodes for local aggregation operations. Their suggested method in conjunction with a cosine-similarity based device filtering attains higher convergence speeds using less number of local updates. Then \cite{ng2021reputation} experiment with FL using clustered devices where the Cluster Heads, selected from the devices, are awarded reputation scores by the members and may communicate with their one-hop neighbors. The results indicate that the proposed configuration improves network efficiency.

The work in \cite{singh2019detailed} compares the communication efficiency of split learning and FL. This research evaluates the learning techniques for increasing client population as well as increasing the size of the global dataset. The results indicate that split learning favors the former whereas the latter is better served by FL. In \cite{nilsson2018performance}, the authors compare FedAvg, Federated Stochastic Variance Reduced Gradient (FSVRG) and CO-OP FL algorithms to investigate the impact of non-IID distribution of various FL optimization techniques. However, the comparison only considers centralized FL aggregation for the mentioned optimization schemes. The results indicate that FedAvg fares better than the other two techniques for non-IID data distribution. A detailed empirical analysis of federated and gossip learning conducted in \cite{hegedHus2021decentralized} shows that gossip performs competitively with federated learning.

 The aforementioned and most remaining body of FL works only use centralized FL as baseline. It therefore, remains to be seen where different FL algorithms may show similar or better performance while being more efficient in other aspects such as communication, energy etc. 

\subsection{Existing Frameworks}
Benchmarking FL takes on greater significance as the envisioned operating environment is highly diverse and different from centralized or silo-based distributed learning. It is therefore necessary that an effective benchmark be capable of allowing heterogeneity in device behavior, data distribution and connectivity. Additionally, the benchmark should be able to support different cooperation schemes that form the basis of different FL algorithms.

Various benchmarking frameworks 
have 
taken different approaches to satisfy the aforementioned objectives. These have been aimed at achieving rapid prototyping \cite{beutel2020flower}, higher scalability \cite{tff, lai2021fedscale} and  realistic data heterogeneity \cite{caldas2018leaf, li2021federated}. Among these options, FedML \cite{he2020fedml} additionally offers topology management by allowing configuration of multiple FL algorithms. However, it is pertinent to note that these frameworks are designed to simulate either FedAvg or a single FL algorithm for a given environment. This indicates that greater work is involved in simulating various FL algorithms to capture more accurate performance details.

\begin{figure}[t]
    \centering
    \includegraphics[scale = 0.45]{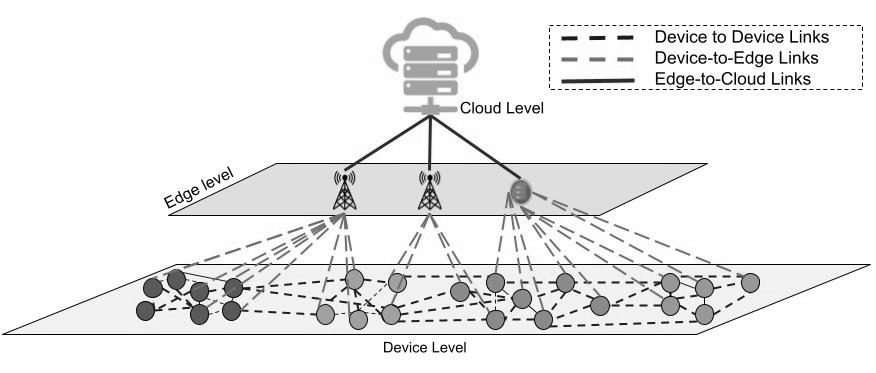}
    \caption{Network topology including Edge network levels and communication links.}
    \label{fig:networklayers}
\end{figure}

This work presents the performance of key FL algorithms distinguished in their ability to utilize various network levels (Fig. \ref{fig:networklayers}). In order to achieve nuanced evaluation and ensure \textit{extensibility}, this work presents the \textbf{FLAGS} framework. The framework allows each network entity, whether a node or a server, to separately emulate a realistic behavior. By implementing a range of functionality associated with each entity, this framework allows diverse network interactions enabling multiple FL algorithms to be simulated easily. Each device has a wide array of associated functions to enable multiple operating conditions and device behaviors. During the experiments, the FL algorithms are subjected to some of the key challenges facing general FL research. The empirical analysis of the performance of these FL algorithms is conducted by varying the operating parameters such as device participation, communication noise and data distribution. 
\section{System Model} \label{sec:sysmodel}
Federated Learning offers an efficient means of 
distributed learning at the Edge Network.  
This section first establishes a \textit{System Model} for the interaction between network entities such as nodes and servers connected by various communication links. The mathematical framework behind the FedAvg algorithm is also presented which is extended in subsequent sections to describe major FL algorithms.

\subsection{Network Topology}
Each of the FL algorithms in this work is evaluated over the same system model to ensure uniformity of conditions. 
The system model is defined using a set of nodes, edge servers and a global cloud server. A stochastic formulation is used to establish the interaction between the nodes at the device layer as well as between the device and hierarchical servers at the edge layer. A set of $\bm{N}$ static nodes are randomly distributed in $\bm{C}$ geo-proximal clusters $(\mathcal{C})$ representing natural grouping. Each cluster may be associated with its own Cluster Head randomly selected from the cluster. Cluster Head selection and Cluster membership are an active area of research, however, beyond the scope of this paper. However, the behavior in the presence of one has been replicated. One node from each cluster $\mathcal{C}_i$ is designated at random as a Cluster Head. This work intends to evaluate the performance under Clustered FL operation and thus assumes automatic membership of proximal nodes.

Each node $k$ is assumed to have a degree $n_k$ and possesses knowledge about its neighborhood. However, it is pertinent to note that a device's neighborhood is not restricted to its cluster and devices can be connected with those in other clusters. This ensures that the network topology at the device layer is \textit{reachable} without involving an entity from a higher edge layer. This layout forms an undirected graph $\mathcal{G} = (\bm{N}, \mathcal{\xi})$ with $\xi$ representing device links or graph edges at the device layer. The links density at the device level is controlled by a set of two parameters $(\gamma, \upsilon)$. 

\begin{table}[!t]
\centering
\begin{tabular}{cl}
\toprule
\textbf{Param}     & \multicolumn{1}{c}{\textbf{Description}}    \\ \midrule
$\bm{\theta^{t}}$    & Global model parameters at time $t$         \\
$\bm{\theta^{t}_k}$ & Model parameters of node $k$ at time $t$    \\
$\bm{\theta^*}$     & Optimal parameters of the global model       \\
$\bm{\mathcal{L}}$  & Global loss function                        \\
$\nabla \mathcal{L}$& Gradient of the loss                        \\
$\bm{x}, \bm{y}$    & Input feature vector, reference output      \\
$\alpha_k$          & Learning rate at node $k$                       \\
$\eta_k$            & Aggregation weight associated with node $k$ \\
$\mathcal{D}$       & Global dataset partitioned among the nodes \\
$\mathcal{D}_k$     & Dataset associated with node $k$            \\
$\bm{N}$       & Number of nodes in the network              \\
$l_k$               & Local loss function associated with node $k$ \\
$\xi_{l,m}$         & Edge link between nodes $l$ and $m$      \\
$\mathcal{C}_i$       & Cluster Head (CH) of the cluster $i$ \\
$\bm{C}$            & Number of geo-proximal clusters in the network\\
$d_k$               & Neighbor cooperation probability of node $k$ \\ 
$p_k$               & Edge server cooperation probability of node $k$ \\ 
\bottomrule
\caption{List of parameters used in the system model and the algorithmic description.
}
\label{tab:params}
\end{tabular}
\end{table}

The probability of a link between geographically proximal devices lying within a cluster $\mathcal{C}_i$ is: $$\underset{m, n \in \mathcal{C}_i}{P(\xi_{mn})} = \{\gamma \ |  \ 0 < \gamma < 1\}$$ 

Similarly, the probability of a device-level link between two nodes belonging to distinct clusters $\mathcal{C}_i$ and $\mathcal{C}_j$ is:
$$\underset{p \in \mathcal{C}_i,q \in \mathcal{C}_j}{P(\xi_{pq})} = \{ \upsilon \ | \ 0 < \upsilon \leq \gamma\}$$ The parameters $\upsilon$ and $\gamma$ are used to alter the network topology from a random placement to a highly clustered setting depending upon the target environment. For these parameters, a setting of ($\gamma \rightarrow 1,\ \upsilon << \gamma )$  indicates that the proximal groups are densely connected with almost each device sharing a link with the other within the group. On the other hand, only a few edges exist between devices between different groups indicating a reduced probability of the devices' ability to communicate directly with devices that are farther off. In contrast, $(\gamma \rightarrow 1, \upsilon \rightarrow \gamma)$ indicates a densely connected device layer where each device is connected to a large number of other devices. Finally, $(\gamma \rightarrow 1, \upsilon = 0)$ indicates that devices only maintain connections within a proximal group without any inter-group connectivity.


The framework can simulate both ideal and noisy links with zero-mean Gaussian noise ($\mathcal{N}(0, \sigma))$. 
Each device has a probability $d_k$ of participating in a neighborhood aggregation and $p_k$ for an edge or cloud server-based aggregation. These probabilities reflect the device status (active/inactive) as well stragglers that may choose not participate in certain aggregation rounds. 
Furthermore, the nodes are assumed to be fully capable of undertaking local learning operations including adjusting hyperparameters. 

\subsection{Federated Learning Methodology}

The global objective in a Federated Learning setting is a weighted sum of the local loss functions:
\begin{equation}\label{globalloss}
    \bm{\mathcal{L}}(\bm{\theta^t}) = \sum_{k=1}^{\bm{N}} \eta_k \mathcal{L}_k(\bm{\theta_k^t}) 
\end{equation}
where $\bm{\mathcal{L}}(\bm{\theta}^t)$ represents the global loss function  and $\mathcal{L}_k(\bm{\theta_k})$ and $\eta_k$ are the local loss function and weight associated with node $k$ at time $t$.
The FL objective is to minimize the weighted sum of the individual loses of $\bm{N}$ devices 
\begin{equation}\label{globalobj}
    \min_{\bm{\theta} \in \mathbb{R}^M} \bm{\mathcal{L}}(\bm{\theta^t}) = \min_{\bm{\theta}} \sum_{k=1}^{\bm{N}} \eta_k \mathcal{L}_k(\bm{\theta_k^t}) 
\end{equation}
Each $\mathcal{L}_k(\bm{\theta_k^t})$ at time $t$ is associated with an $M$-dimensional model parameterized by $\bm{\theta_k^t} \in \mathbb{R}^M$ and  the weight $\eta_k$ for node $k$. The initial parameters of the model, $\bm{\theta}^t$, are shared with the devices through a server. The local loss at a device is calculated through the training as follows 
\begin{equation}\label{localobj}
\mathcal{L}_k(\boldsymbol{\theta_k^t}) = \dfrac{1}{|\mathcal{B}_k|} \sum\limits_{(\boldsymbol{x_i}, \boldsymbol{y_i}) \in \mathcal{B}_k} l(\boldsymbol{x_i}, \boldsymbol{y_i}; \boldsymbol{\theta}_k^{t})
\end{equation}
where $l(\mathbf{x}_i, \mathbf{y}_i; \boldsymbol{\theta_k^t})$ represents the loss of the machine learning task (e.g cross-entropy, mean squared error) calculated for the local model for the current model parameters. This is calculated for each point $(\bm{x_i}, \bm{y_i})$ in the minibatch $\mathcal{B}_k$ sampled from the local dataset $\mathcal{D}_k$. Using the principles of Distributed Stochastic Gradient Descent (DSGD), each device with the learning rate $\alpha_{kr}$ for a round $r$, updates its model as:
\begin{equation}\label{updatedmodel}
    \bm{\theta}_k^{t+1} = \boldsymbol{\theta}_k^t - \alpha_k^t \nabla \mathcal{L}_k^t(x_i, y_i; \boldsymbol{\theta}_k^t)
\end{equation}
where $\nabla \mathcal{L}_k^t(x_i, y_i; \boldsymbol{\theta}_k^t)$ are the gradients calculated by node $k$ during the local update round $t$ and $\alpha^t_k$ is the learning rate associated with $k$th node at time $t$.
The updated model parameters from Eq. \ref{updatedmodel} or the gradients calculated above are then shared with the server. The global model is obtained through the weighted aggregation of the model parameters
\begin{equation}\label{modelaggregate}
\bm{\theta}^{t+1} = \bm{\theta}^t +  \sum \limits_{k=1}^{\bm{N}} \eta_k \bm{\theta}_k^t   
\end{equation}
For an alternate FL scheme employing gradient sharing, Eq. \ref{modelaggregate} can be formulated as:
\begin{equation}\label{gradaggregate}
    \bm{\theta}^{t+1} = \bm{\theta}^t - \alpha^t \sum\limits_{k=1}^{\bm{N}} \eta_k \nabla \mathcal{L}_k^t (\bm{x}_i, \bm{y}_i ; \bm{\theta}_k^t)
\end{equation}
where $\bm{\theta}^t$ refers to the parameters of the global model used at the start of round $t$ and $\alpha^t$ is the learning rate employed by the server. The learning rate $\alpha_t$ is typically kept smaller \cite{konevcny2016federated} compared with centralized learning on large datasets. 
As evident from Eq. \ref{globalobj}-\ref{gradaggregate}, the entire Federated Learning process bypasses the need for the devices to share data at any stage, offering a strong privacy advantage over centralized forms of ML. It instead relies on the communicating model parameters indicated in Eq. \ref{modelaggregate} and Eq. \ref{gradaggregate} to the local server. The server is also required to share the aggregated model with the connected nodes for the subsequent rounds adding to the overall communication volume needed to execute FL successfully. Such servers are typically located at the network cloud level, offering multiple services. Thus, frequent communication to and from these servers becomes an extremely expensive operation in addition to straining the latency requirements of various edge services in addition to subjecting the shared models to additional adversarial risks.Table~\ ref{tab:params} presents the list of parameters used in this work.

\section{FLAGS Framework} \label{sec:flags}
The FL algorithms are envisioned for a highly heterogeneous setting with limited prior information. 
Accurate assessment of their performance requires that the simulation framework provide repeatable and uniform conditions, support easily configurable FL algorithms, possess flexibility to implement new ones and cater to multiple types of communication links. This work develops \textbf{F}ederated \textbf{L}earning \textbf{A}l\textbf{G}orithms \textbf{S}imulation (\textbf{FLAGS}) framework with the intent of making it easier to configure multiple FL algorithms under a wide range of operating conditions.
\begin{figure}[!t]
    \centering
    \includegraphics[scale = 0.2]{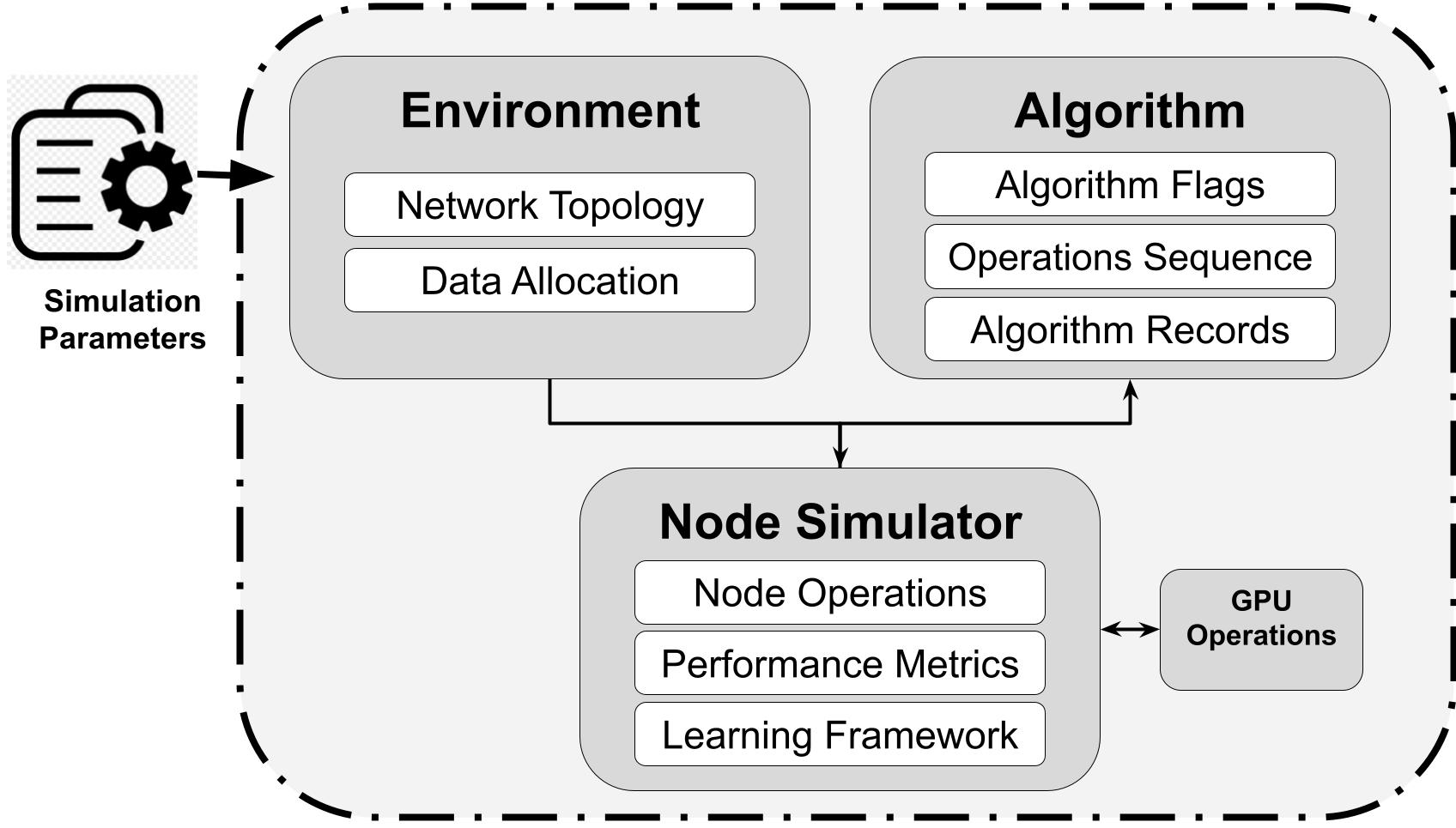}
    \caption{FLAGS Framework Architecture}
    \label{fig:framework}
\end{figure}
\subsection{FLAGS: Features}
FLAGS is a lightweight FL prototyping framework allowing for an accurate assessment of multiple FL algorithms for highly realistic network topologies.
Inherent in FLAGS are different levels of device participation (controlled by parameters $p_k$ for participation in central aggregation and $d_k$ for controlling neighborhood aggregation), device selection mechanism for nodes and servers as well as generating multiple data distributions including IID, non-IID and extremely-skewed non-IID data partitions. It also supports synchronous and deadline-based asynchronous operation to replicate heterogeneous device behavior where devices training progresses at different paces. The highlight of FLAGS lies in its capability to configure and subject multiple FL algorithms to a realistic multi-tiered environment which allows users to propose the best environment-FL fit. 

\subsection{FLAGS Architecture}
FLAGS is developed so that multiple FL algorithms may be simulated using a single framework and tested under a unified set of conditions. The FLAGS architecture is highly modular and uses three different functional blocks to control the respective set of operations, as depicted in Fig. \ref{fig:framework}.
These are the \textit{Environment, Algorithm} and \textit{Node} modules. 

The \textit{Environment} block is responsible for configuring the network topology for FL operations. It controls the generation of the network topology and the respective links while providing access to related information. Once configured, this module enables interaction within various entities of the network. The device layer is configured with the  devices grouped into $\bm{C}$ proximal cluster. Each of the devices is also connected to the respective server at the Edge Layer. These hierarchical servers, in turn, are then connected to a global server representing the cloud level. The overall configuration in the environment module affords greater flexibility in conjunction with the corresponding \textit{flags}. This feature allows multiple FL algorithms to be simulated by just changing the \textit{flags} configuration. This distinguishes the \textit{FLAGS} framework from others since those have been designed to simulate a single topology.

\begin{table}[t]
\centering
\begin{tabular}{m{0.12\textwidth}P{0.19\textwidth}P{0.19\textwidth}P{0.19\textwidth}P{0.2\textwidth}}
\toprule
\textbf{Flags } & \textbf{D2D Aggregation} & \textbf{Edge Aggregation} & \textbf{Cluster Aggregation} & \textbf{Inter-Cluster Aggregation} \\ \midrule
\textbf{FedAvg}              & 'CServer'                   & False                     & False                        & False                              \\
\textbf{HFL}                 & False                       & True                      & False                        & False                              \\
\textbf{D2DFL}               & 'D2D'                       & False                     & False                        & False                              \\
\textbf{GFL}                 & 'Random'                    & False                     & False                        & False                              \\
\textbf{HD2DFL}              & 'D2D'                       & True                      & False                        & False                              \\
\textbf{HGFL}                & 'Random'                    & True                      & False                        & False                              \\
\textbf{CFL}                 & False                       & False                     & True                         & False                              \\
\textbf{iCFL}                & False                       & False                     & True                         & True                               \\
\textbf{iCD2DFL}             & 'D2D'                       & False                     & True                         & True\\
\bottomrule
\end{tabular}\caption{FL Algorithms with the respective \textit{flag} settings for the \textbf{FLAGS} framework}\label{table:flagsettings}
\end{table}

The \textit{Algorithm} module manages the configuration and sequence of operations for all FL algorithms. It makes use of a set of \textit{flags} to generate unique interaction between various elements of the environment. FLAGS implements four different flags to generate the required interaction:
\begin{enumerate*}[label = (\alph*)]
\item Device Aggregation
\item Edge Aggregation
\item Cluster Aggregation and
\item Inter-Cluster Aggregation.
\end{enumerate*}
The Device Aggregation flags control whether a node is allowed to share information with its peers or not. This may be within the devices' own neighborhoods or through random pairings including multi-hop interactions. The Edge Aggregation, when enabled, enables the devices to share their local models with the hierarchical servers. Cluster and Inter-Cluster Aggregations have been included in the framework to support expandability for FL scenarios with device clusters.
Employing a flag-based mechanism allows the framework to remain configurable for addition of more FL algorithms.The \textit{flag} settings for the various FL algorithms covered within this work have been shared under Table~\ref{table:flagsettings}.  

Finally, the \textit{Node} block implements all device level operations including training, testing and communication while maintaining records all performance metrics. This module enables the nodes to communicate with both Environment and Algorithm blocks to obtain the required topological and permissible-interaction information. The \textit{devices} and \textit{servers} both are generated using this block imbuing them with a range of functionality available to modern devices for conducting the learning operations.

\begin{figure*}[t]
\centering
\def\twidth{0.6}
\subfloat[HFL]{%
\includegraphics[scale = 0.15]{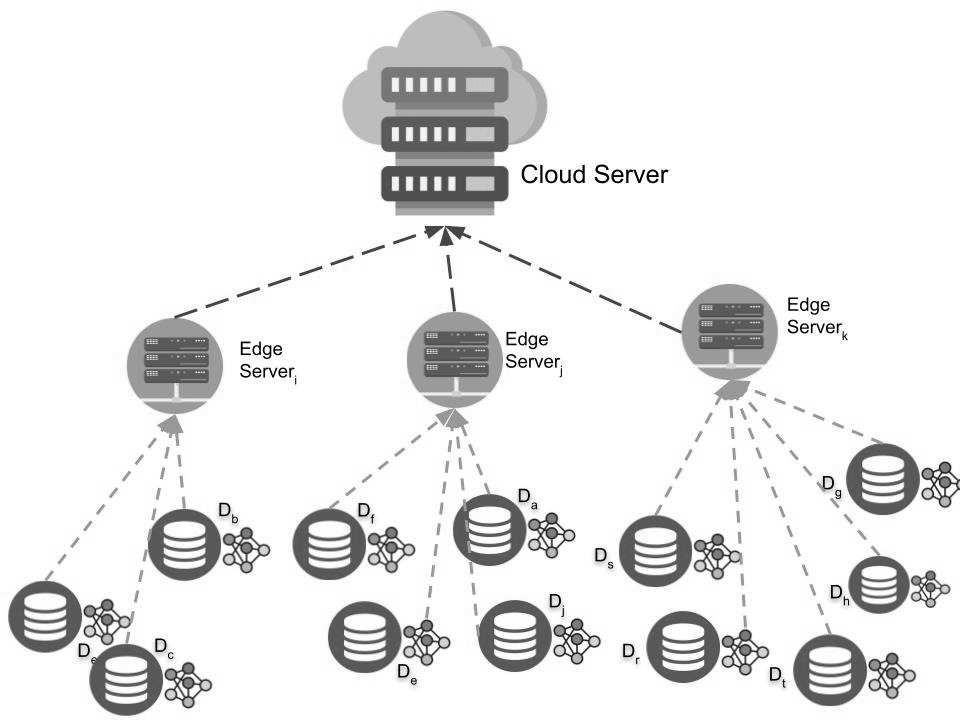}%
}\hfill
\subfloat[D2DFL]{%
\includegraphics[scale =0.15]{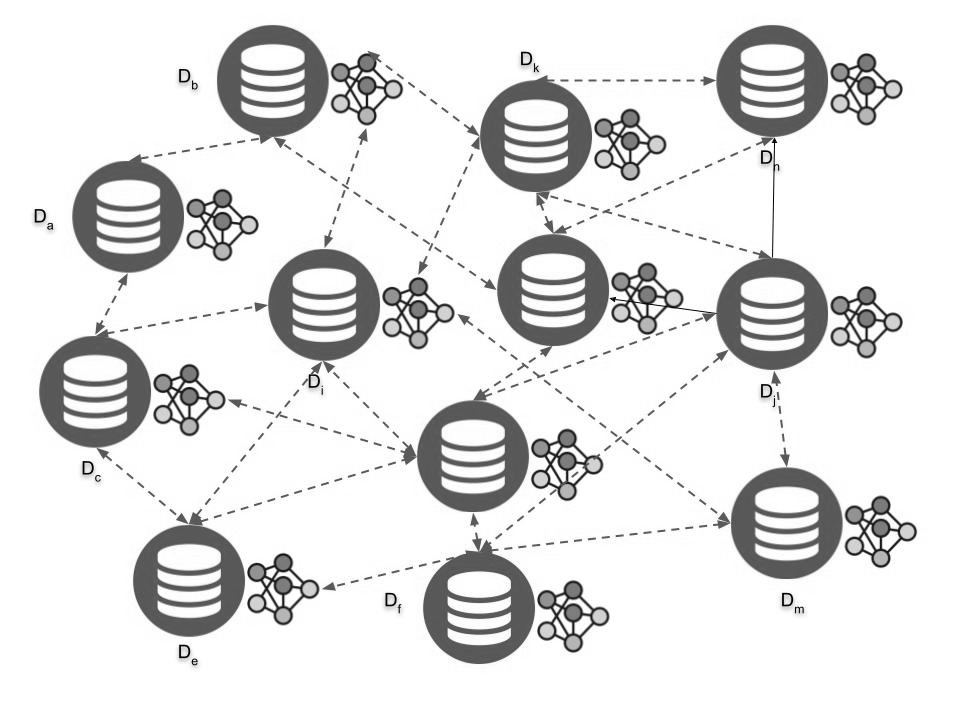}%
}
\subfloat[GFL]{%
\includegraphics[scale = 0.15]{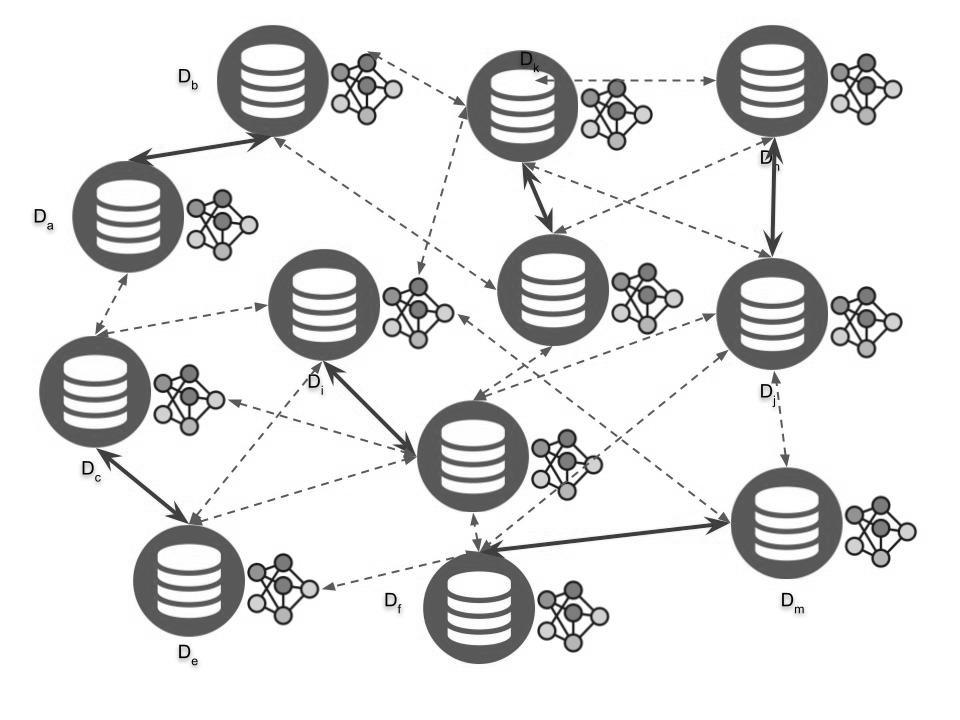}%
}\hfill
\caption{Main Federated Learning Algorithms. Dashed lines represent links between various entities whereas only solid lines in (c) depict active established-pair links in GFL. }
\label{fig:FLtopo}
\end{figure*}

\section{Federated Learning Algorithms}\label{sec:FLclasses}
Federated Learning was initially conceived as a two-tiered learning framework alternating between clients/nodes performing the training and the cloud servers aggregating the local models to yield a global model. In order to increase the efficiency, the envisaged FL was further spread over the Edge Network to benefit not only from the Edge devices capabilities but also address the challenges inherent with centralization. This evolution has yielded various FL algorithms suitable to a variety of scenarios including vehicular-networks, dense IoT networks, cross-silo operations between relatively large data centers. These algorithms employ three different types of communication links as shown in Fig. \ref{fig:networklayers} associated with different communication costs:
\begin{enumerate*}[label = (\alph*)]
    \item Device-to-Device (D2D) 
    \item Device-to-Edge (D2E)
    \item Edge-to-Cloud (E2C) 
\end{enumerate*}

D2D link are limited to the device level whereas D2E links connect the nodes with the edge servers. Finally, E2C links connect the edge servers to the core network. While D2D links and D2E links operate independently, E2C links require the latter for information to be transmitted from the devices to the edge server for onward transmission to the core network. All subsequent communication analyses in this work use this fact to establish effectiveness of a particular form of communication in FL operations.

\begin{algorithm}[t]
\DontPrintSemicolon
  \KwInput{Initial Model parameters $\bm{\theta}_0$, $\mathcal{S}$ hierarchical servers  with device association}
  \KwOutput{Global model parameters $\bm{\theta}_{t\rightarrow \infty}$}
  Initialization by global server: $\bm{\theta}_0$ received by each hierarchical server $S_i$ and broadcast among the connected nodes \;
  \For{Nodes $k\ =\ 1\ \dots\dots \bm{N}$ (\textbf{in parallel})}{
  Perform Local update for $e$ epochs\;
  $\bm{\theta_{k}^{t+\tau}} = \bm{\theta}_{k}^t - \alpha_{k}^t \nabla \mathcal{L}_k(x_{m}, y_{m}; \bm{\theta}_{k}^t)$ \;
  \textbf{TX: } With a probability $p_k$, send the $\bm{\theta}_k^{t+\tau}$ to the hierarchical server for aggregation\;
  }
  \For{Servers  $\mathcal{S}_i \ \forall \ i =\ 1\ \dots\dots \mathcal{K}$  (\textbf{in parallel})}{
  \textbf{AGG: }Aggregate the models received from the $n_i$ nodes \;
  $\bm{\theta}_{S_i}^{t+T} = \bm{\theta}_{S_i}^t + \sum\limits_{j \in n_i} w_{j} \bm{\theta}_j^{t + \tau}$ 
  }
  \If{round  \% f == 0}
    {
        \For{Servers $S_i  \ \forall \  i = 1 \dots\dots \mathcal{K}$  (\textbf{in parallel})}{
        Upload models to the global server \;
        }
        \textbf{Global Aggregation} \;
        $\bm{\theta}^{t+1} = \bm{\theta}^t
        +  \sum\limits _{j \in \mathcal{K}} \eta_j \bm{\theta}_j^{t + T} $ \;
        Send model back to Edge Servers $S_i$ \;
        Edge Servers $S_i$ broadcast model parameters $\bm{\theta}^{t+1}$ to all nodes connected respectively \;
    }
\caption{Hierarchical Federated Learning} \label{HFL}
\end{algorithm}

A key aspect in major FL literature is the absence of cross-configuration comparison and evaluation of FL algorithms. Such an analysis promises to identify optimal employment of these FL algorithms based on the operating environment and nature of the problem being addressed. This section first provides a description of these algorithms followed by details of the experiments in the subsequent sections.  

\subsection{Hierarchical Federated Learning} Hierarchical Federated Learning (HFL) introduces intermediate model aggregation closer to the data origin using an edge server \cite{liu2020client}. This FL algorithm introduces $\mathcal{F}$ aggregation layers between the nodes and the global cloud server, $\mathcal{F}$ increasing toward the cloud layer. Cellular base stations or Mobile Edge Computing (MEC) Servers are envisioned to realize this role \cite{Wang_Han_Leung_Niyato_Yan_Chen_2020} for their respective connected devices. Each network level, as depicted in Fig. \ref{fig:FLtopo}(a), aggregates the models received from the previous layer and passes them to the next layer in the hierarchy. This scheme offers reduction in the overall upstream communication as well progressively offloading the computational load at the edge network. 

Each hierarchical layer $\mathcal{F}$ houses $\mathcal{K}_{\mathcal{F}}$  edge servers. Each server at layer $\mathcal{F} = 1$  aggregates models from the devices linked to it respectively. The pseudocode for HFL have been laid out in Alg. \ref{HFL}. The initial model parameters i.e. $\bm{\theta}_0$ are initialized by the global server $\mathcal{S}_g$ and shared with the nodes. The nodes in turn update the model over a mini-batch of size $\mathcal{M}$ to yield $\bm{\theta}_k^{t+1}$. After completing $e$ local updates, each node shares its updated parameters with the hierarchical server $S_i$ with a probability $p_k$. The servers aggregate the received models which is then shared with the servers in the next layer/global server after $f$ aggregation rounds. At this stage, the global server aggregates the input models from the $\mathcal{K}$ hierarchical servers and generates the global model $\bm{\theta}^{t+\tau}$ which are then shared through the local servers with the associated nodes.\\

\begin{algorithm}[t]
\DontPrintSemicolon
    \KwInput{Model parameters $\theta_k$ for each node, degree $n_k$}
    \KwOutput{D2D Aggregated Models}
    \For{Nodes $k\ =\ 1\ \dots\dots \bm{N}$ \textbf{in parallel}}{
    Update local model for $e$ epochs \;
    $\bm{\theta_{k}^{t+\tau}} = \bm{\theta}_{k}^t - \alpha_{k}^t \nabla \mathcal{L}_k(x_m, y_m; \bm{\theta}_{k}^t)$ \;
    Send model to 1-hop neighbors $n_k$ with probability $d_k$ \;
    Receive models from $n_k$ 1-hop neighbors \;
    Perform weighted aggregation for the received models \;
    $\bm{\theta}^{t+1}_k = \bm{\theta}^{t + \tau}_k
        +  \sum\limits _{j \in n_k} \eta_j \bm{\theta}_j^{t +\tau} $ \;
    }
\caption{Device-to-Device Federated Learning}\label{D2DFL}
\end{algorithm}

\subsection{Device-to-Device Federated Learning}
With Device-to-Device (D2D) communication \cite{tehrani2014device}, FL can be operated without the requirement of a central aggregator \cite{Savazzi_Nicoli_Rampa_2020}. In a purely serverless FL framework, the nodes communicate with their immediate neighbors and resort to local aggregation Fig. \ref{fig:FLtopo}(b). At a given time $t$, the nodes share their locally learned parameters $\bm{\theta}_k^t$ with their neighborhood $n_k$. 
The recipients in turn perform aggregation of the received models combining the incoming parameters $\bm{\theta}_k^t$ with corresponding weights $\{\eta_j \forall j \in n_k\}$ to generate $\bm{\theta}_k^{t+1}$:
The entire operation of D2DFL is depicted in Alg.~\ref{D2DFL}. \\



\begin{algorithm}[H]
\DontPrintSemicolon
    \KwInput{Model parameters $\bm{\theta}_k$, optimization parameters}
    \KwOutput{Gossip-aggregated model}
    \For{Nodes  $k\ =\ 1\ \dots\dots \bm{N}$ \textbf{in parallel}}{
    $\bm{\theta}_k^{t+\tau} \longleftarrow$ Model-Update($\bm{\theta}_k^(t)$) \;
    $\bm{\theta}_k^{t+ 1} \longleftarrow Gossip \ Exchange \ (\theta^{t+\tau}_k)$
    }
    \SetKwFunction{FUpdate}{Model-Update}
    \SetKwFunction{GossipEx}{Gossip Aggregate}
    
    \SetKwProg{Fn}{Function}{:}{}
    \Fn{\FUpdate{$\bm{\theta}_l^t$}}{
    \For{batch $b$ of size $\mathcal{M} \in \mathcal{D}_l $}{
    $\xi_b = (x_b, y_b)$ is the data sample pairs in $b$\;
    $\bm{\theta_{l}^{t+\tau}} = \bm{\theta}_{l}^t - \alpha_{}^t \nabla \mathcal{L}_l(\xi_b; \bm{\theta}_{l}^t)$ \;
    }
    \KwRet $\bm{\theta}_l^{t+\tau}$\;
    }
    \Fn{\GossipEx{$\bm{\theta}_l^t$}}{
    Perform aggregation handshake with a random node \;
    Exchange models for aggregation between the pair \;
    Aggregate received model \;
    $\bm{\theta}^{t+T}_l = \bm{\theta}^{t+ \tau}_l
        +  \eta_j \bm{\theta}_j^{t + \tau} $ \;
    \KwRet $\bm{\theta}^{t+T}_i$
    }
\caption{Gossip Federated Learning}\label{GFL}
\end{algorithm}

\subsection{Gossip Federated Learning}
Gossip communication is another form of decentralized communication wherein the nodes communicate with randomly selected peer nodes. Gossip Federated Learning (GFL)  was among the early fully decentralized FL frameworks to have been proposed \cite{daily2018gossipgrad}. For a network with $\bm{N}$ nodes, the nodes form random pairs with other nodes from the network forming a gossip pair $(i,\ j\ \forall\ \  i,\ j \in \ \bm{N})$ as shown in Fig. \ref{fig:FLtopo}(c). Once the connection has been established, the pair exchange their locally updated models $\bm{\theta}_i^t$,  $\bm{\theta}_j^t$ and in turn aggregate with the received model parameters to generate $\bm{\theta}_i^{t+1}$ and $\bm{\theta}_j^{t+1}$. The process results in considerably curtailed communication costs. The details of the operation have been provided in Alg.~\ref{GFL}. 
\subsection{Hierarchical Device-to-Device Federated Learning}
Hierarchical D2DFL employs aggregation 
at the device levels, edge servers and cloud servers. Thus
the D2D, D2E and E2C links all are employed to enable such operation
\cite{Lin_Hosseinalipour_Azam_Brinton_Michelusi_2021}. 
The proposed mechanism allows for devices to perform local updates on their model parameters $\mathcal{\bm{\theta}}_k^t$ and share them with their $n_k$ neighbors leading to device level aggregation. Simultaneously, the devices also share the model parameters with the connected servers $\mathcal{S}_i$. The servers in turn aggregate the received models and share along the hierarchy $\mathcal{F}$ eventually leading up to the global server $\mathcal{S}_g$ as indicated in Fig. \ref{fig:FLtopo}(d). The global model $\bm{\mathcal{\theta}}^{t+\tau}$ gets disseminated back to the nodes through the respective servers and the process is continued till convergence. The details of the entire process has been outlined in Alg. \ref{HD2DFL}. While greater cooperation is envisioned with this algorithm, the learning requires communication at both device and upstream level.
\subsection{Hierarchical Gossip Federated Learning}
Hierarchical FL, as described in Alg.\ref{HFL}, allows device layer to interact with Edge Servers to aggregate local models resulting in sub-global models. These models are then passed onto global server for generating the global model $\bm{\theta}^t$. HD2DFL builds on this configuration by allowing devices layer entities to interact among themselves before sharing the local aggregated models with the higher edge layers. The performance as shall be seen in Sec-\ref{sec:analysis} indicate better performance albeit at the communication volume which is almost the joint sum of HFL and D2DFL. This observation led the researchers in \cite{hashemi2021benefits} to introduce gossip steps at the device level instead of the full neighborhood aggregation as indicated in Alg \ref{alg:HGFL}. \\

\begin{algorithm}[H]
\DontPrintSemicolon
  \KwInput{Initial Model parameters $\bm{\theta}_0$, Neighborhood information $\mathcal{G}_k = \{\bm{N}, \mathcal{\xi}_k \}$, Hierarchical servers $K$ with device association}
  \KwOutput{Global model parameters $\bm{\theta}_{t\rightarrow \infty}$}
  Initialization by global server: $\bm{\theta}_0$ received by each hierarchical server $S_i$ and broadcast among the connected nodes $n_i$\;
  \For{Node $k\ =\ 1\ \dots\dots \bm{N}$ \textbf{in parallel}}{
    $\bm{\theta}_k^{t+\tau} \longleftarrow Model \ Update (\bm{\theta_k^t})$ \;
    $\bm{\theta}^{t+\delta}_k \longleftarrow \ D2D \ Aggregate (\bm{\theta_k^{t +\tau}}, n_k)$ \ ;    }
  \If{aggregation round \% f == 0}{
  \For{Servers $S_i  \ \forall \  i = 1 \dots\dots H$  (\textbf{in parallel}}{
  Sample $q$ from the associated nodes and request parameters \;
  Aggregate received parameters: $\bm{\theta}_{S_i}^{t+\tau} = \bm{\theta}_{S_i}^t + \sum\limits_{j \in q} w_{j} \bm{\theta}_j^t$}
    {
        \For{Servers $S_i  \ \forall \  i = 1 \dots\dots H$  \textbf{in parallel}}{
        Upload model $\bm{\theta}_{S_i}^{t}$ to the global server
        }
        \textbf{Global Aggregation} : $\bm{\theta}_{t+1} = \bm{\theta}_t
        +  \sum\limits _{j \in S_i} \eta_j \bm{\theta}_j^t $ \;
        Send model back to nodes through respective Edge Servers $S_i$
        }
    }

  \SetKwFunction{FUpdate}{Model-Update}
  \SetKwFunction{SAggregate}{Aggregate}
    \SetKwFunction{DDAgg}{D2D-Aggregate}
    
      \SetKwProg{Fn}{Function}{:}{}
      \Fn{\FUpdate{$\bm{\theta}_l^t$}}{
        \For{batch $b$ of size $\mathcal{M} \in \mathcal{D}_l $}{
        $\xi_b = (x_b, y_b)$ is the data sample pairs in $b$\;
        $\bm{\theta_{l}^{t+\tau}} = \bm{\theta}_{l}^t - \alpha_{l}^t \nabla \mathcal{L}_l(\xi_b; \bm{\theta}_{l}^t)$ \;
        }
        \KwRet $\bm{\theta}_l^{t+\tau}$\;
        }
    
    \Fn{\DDAgg{$\bm{\theta}_k^t,\ n_l$}}{
        Exchange model with 1-hop neighbors $n_l$ with probability $d_l$ \;
        Perform weighted aggregation for the received models \;
        $\bm{\theta}^{t+\delta}_l = \bm{\theta}^{t + \tau}_l
            +  \sum\limits _{j \in n_l} \eta_j \bm{\theta}_j^{t +\tau} $ \;
        \KwRet $\bm{\theta}^{t+\delta}_l$
            }  
\caption{Hierarchical-D2D Federated Learning}\label{HD2DFL}
\end{algorithm}

The nodes communicate at the device level using random pairings resulting in gossip communication. These random node pairs exchange local models during this stage to perform local aggregation yielding $\bm{\theta}^{t+\tau}_k$ for $k = 1 \dots \dots \bm{N}$ network nodes. Subsequently, these gossip-aggregated models are shared with the edge servers for hierarchical aggregation and subsequently cloud aggregation respectively.\\

\begin{algorithm}[H]
\DontPrintSemicolon

  \KwInput{Initial Model parameters $\bm{\theta}_0$, Neighborhood information $\mathcal{G}_k = \{\bm{N}, \mathcal{\xi}_k \}$, Hierarchical servers $K$ with device association}
  \KwOutput{Global model parameters $\bm{\theta}_{t\rightarrow \infty}$}
  Initialization by global server: $\bm{\theta}_0$ received by each hierarchical server $S_i$ and broadcast among the connected nodes $n_i$\;
  \For{Node $k\ =\ 1\ \dots\dots \bm{N}$ \textbf{in parallel}}{
    $\bm{\theta}_k^{t+\tau} \longleftarrow Model \ Update \ (\bm{\theta}_k^{t})$ \;
    $\bm{\theta}^{t+\delta}_k \longleftarrow Gossip \  Exchange \ (\bm{\theta}_k^{t+\tau})$ \;
    }
  \If{aggregation round \% f == 0}{
  \For{Servers $S_i  \ \forall \  i = 1 \dots\dots H$  (\textbf{in parallel}}{
  Sample $q$ from the associated nodes \;
  \For{$j \in$ Sampled Nodes}{Share model parameters $\bm{\theta}_j^{t + \delta}$ with the server
    }
  Aggregate the received parameters \;
  $\bm{\theta}^{t+T}_{S_i} = \bm{\theta}_{S_i}^{t+\delta} + \sum\limits_{j \in q} w_{j} \bm{\theta}_j^{t+\delta}$   }
    {
        \For{Servers $S_i  \ \forall \  i = 1 \dots\dots H$  (\textbf{in parallel}}{
        Upload model $\bm{\theta}_{S_i}^{t+T}$ to the global server \;
        }
        \textbf{Global Aggregation} :   $\bm{\theta}_{t+1} = \bm{\theta}_{t}
        +  \sum\limits _{j \in S_i} \eta_j \bm{\theta}_j^{t+T} $ \;
        Send model back to the nodes through respective Edge Servers $S_i$\;
    }
    }
    \SetKwFunction{FUpdate}{Model-Update}
    \SetKwFunction{GossipEx}{Gossip Aggregate}
      \SetKwProg{Fn}{Function}{:}{}
      \Fn{\FUpdate{$\bm{\theta}_l^t$}}{
        \For{batch $b$ of size $\mathcal{M} \in \mathcal{D}_l $}{
        $\xi_b = (x_b, y_b)$ is the data sample pairs in $b$\;
        $\bm{\theta_{l}^{l+\tau}} = \bm{\theta}_{l}^t - \alpha_{l}^t \nabla \mathcal{L}_l(\xi_b; \bm{\theta}_{l}^t)$ \;
        }
        \KwRet $\bm{\theta}_l^{t+\tau}$\;
        }
    \Fn{\GossipEx{$\bm{\theta}_l^t$}}{
    Perform aggregation handshake with a random node \; 
    Exchange models for aggregation between the pair \;
    Aggregate received model : $\bm{\theta}^{t+\zeta}_l = \bm{\theta}^{t+ \tau}_l + \eta_j \bm{\theta}_j^{t + \tau} $ \;
    \KwRet $\bm{\theta}^{t+\zeta}_l$
    } 
\caption{Hierarchical-Gossip Federated Learning}\label{alg:HGFL}
\end{algorithm}

\subsection{Clustered Federated Learning}
Clustered FL (CFL) builds on the fully decentralized Federated Learning while aiming to reduce communication volume by clustering operations. The nodes in proximal locations are grouped into Clusters formed around a Cluster Head (CH). The devices become a Cluster-Member (CM) by associating with a Cluster Head $\mathcal{C}_i$, itself a device. During the training phase, the device perform local updates of their own models i.e $\bm{\theta}_k^t$. At the aggregation stage, each CM $i$ shares its local model with the CH i.e $\mathcal{C}_i$, with a probability $d_k$. Subsequently, all the received models are aggregated to generate the sub-global model $\bm{\theta}^t_{\mathcal{C}_i}$. The aggregated model is shared with the respective CMs of $i^{th}$ cluster. The process continues until an acceptable threshold is reached. The details of the entire operation are presented in Alg. \ref{alg:CFL}.\\

\begin{algorithm}[H]
\DontPrintSemicolon
    \KwInput{Model parameters $\theta_k$ for each node, degree $n_k$}
    \KwOutput{D2D Aggregated Models}
    Devices $i = 1...... \bm{C}$ chosen as Cluster Head (CH) of cluster $\mathcal{C}_i$\;
    \For{free nodes $k = 1...... \bm{N}-\bm{C}$}{
    Join $\mathcal{C}_i \ \forall \ i \in \bm{C}$ as a Cluster-Member (CM)\;}
    \For{Cluster Head $\mathcal{C}_i \forall \ i = 1 \dots \dots \bm{C}$ in parallel}
    {Send model to CM $\{p \ : \ p \in \mathcal{C}_i\}$ \;
     \For{Nodes $k\ =\ 1\ \dots\dots \bm{N}$ \textbf{in parallel}}{
     $\bm{\theta}^{t + \tau}_k \longleftarrow $ Model-Update ($\bm{\theta}_k^t$) \;
    }
    Receive models from CM $\{p \ : \ p \in \mathcal{C}_i\}$ with probability $d_p$ \;
    Perform weighted aggregation of received models \;
    $\bm{\theta}^{t+1}_{\mathcal{C}_i} = \bm{\theta}^{t+ \tau}_{{\mathcal{C}_i}}
        +  \sum\limits _{j \in C_i} \eta_j \bm{\theta}_j^{t + \tau} $ \;
    }
    \SetKwFunction{FUpdate}{Model-Update}
      \SetKwProg{Fn}{Function}{:}{}
      \Fn{\FUpdate{$\bm{\theta}_l^t$}}{
        \For{batch $b$ of size $\mathcal{M} \in \mathcal{D}_l $}{
        $\xi_b = (x_b, y_b)$ is the data sample pairs in $b$\;
        $\bm{\theta_{l}^{t+\tau}} = \bm{\theta}_{l}^t - \alpha_{l}^t \nabla \mathcal{L}_l(\xi_b; \bm{\theta}_{l}^t)$ \;
        }
        \KwRet $\bm{\theta}_l^{t+\tau}$\;
        }
\caption{Clustered Federated Learning}\label{alg:CFL}
\end{algorithm}

\subsection{Clustered Device-to-Device Federated Learning (CD2DFL)}Clustered D2DFL (CD2DFL) builds on the fully decentralized Federated Learning while aiming to reduce communication volume by clustering operations. The devices join a cluster by associating with a Cluster Head $(CH)_i$, itself a device. The Cluster-Members (CMs) in this case, continue local updates and D2D aggregating models over their respective neighborhoods. After a number of local aggregation rounds, each CH orchestrates a cluster-level aggregation wherein the CMs share their respective models with the CH. Once aggregated, the CMs receive the aggregated model $\bm{\theta}_{\mathcal{C}_i}^{t+1}$ from the CH and continue with the D2DFL. Alg. \ref{alg:CD2DFL} outlines the overall procedure for CD2DFL.\\

\begin{algorithm}[H]
\DontPrintSemicolon
    \KwInput{Model parameters $\theta_k$ for each node, degree $n_k$}
    \KwOutput{D2D Aggregated Models}
    Devices $i = 1...... \bm{C}$ chosen as Cluster Head (CH) of cluster $\mathcal{C}_i$\;
    \For{free nodes $k = 1...... \bm{N}-\bm{C}$}{
    Join $\mathcal{C}_i \ \forall \ i \in \bm{C}$ as a Cluster-Member (CM)\;}
    \For{Cluster Head $\mathcal{C}_i \forall \ i = 1 \dots \dots \bm{C}$ in parallel}
    {Send model to CM $\{p \ : \ p \in \mathcal{C}_i\}$  \;
     \For{Nodes $k\ =\ 1\ \dots\dots \bm{N}$ \textbf{in parallel}}{
     $\bm{\theta}^{t + \tau}_k \longleftarrow $ Model-Update ($\bm{\theta}_k^t$) \;
    $\bm{\theta}^{t+T}_k \longleftarrow $ D2D-Aggregate ($\bm{\theta}_k^{t + \tau}, n_k$)
    }
    Receive models from CM $\{p \ : \ p \in \mathcal{C}_i\}$ with probability $d_p$\;
    Perform weighted aggregation \;
    $\bm{\theta}^{t+1}_{\mathcal{C}_i} = \bm{\theta}^{t+ T}_{{\mathcal{C}_i}}
        +  \sum\limits _{j \in C_i} \eta_j \bm{\theta}_j^{t +T} $
    }
    \SetKwFunction{FUpdate}{Model-Update}
    \SetKwFunction{DDAgg}{D2D-Aggregate}
      \SetKwProg{Fn}{Function}{:}{}
      \Fn{\FUpdate{$\bm{\theta}_l^t$}}{
        \For{batch $b$ of size $\mathcal{M} \in \mathcal{D}_l $}{
        $\xi_b = (x_b, y_b)$ is the data sample pairs in $b$\;
        $\bm{\theta_{l}^{t+\tau}} = \bm{\theta}_{l}^t - \alpha_{l}^t \nabla \mathcal{L}_l(\xi_b; \bm{\theta}_{l}^t)$ \;
        }
        \KwRet $\bm{\theta}_l^{t+\tau}$\;
        }
    
    \Fn{\DDAgg{$\bm{\theta}_l^t,\ n_l$}}{
        Exchange models with 1-hop neighbors $n_l$ with probability $d_l$ \;
        Perform weighted aggregation : $\bm{\theta}^{t+\delta}_l = \bm{\theta}^{t + \tau}_l
            +  \sum\limits _{j \in n_l} \eta_j \bm{\theta}_j^{t +\tau} $ \;
        \KwRet $\bm{\theta}^{t+\delta}_l$\;
        }    
\caption{Clustered Device-to-Device Federated Learning}\label{alg:CD2DFL}
\end{algorithm}

\subsection{Inter-Cluster Device-to-Device Federated Learning}
Clustered FL in the presence of D2D interactions significantly improves local cooperation. In this regard, the devices ability to interact within its neighborhood and subsequent consolidation at Cluster Head (CH) results in a greater participation in the overall aggregation scheme. In order to further expand this functionality, this algorithm allows the Cluster Heads to communicate the other CH's using a gossip mechanism. While the rest of the  learning and aggregation process remains similar to CD2DFL, the aggregation at the cluster head level is followed by the CH exchanging the locally aggregated cluster model $\bm{\theta}_{\mathcal{C}_i}^t$ with other CH $\{\mathcal{C}_j | i \neq j\}$ using randomized gossip. Each CH $\mathcal{C}_i$ is allowed a limited number of gossip steps before sharing the gossip-aggregated model $\bm{\theta}_{\mathcal{C}_i}^{t+\tau}$ with its respective CMs. As shown in Alg. \ref{alg:iCD2DFL}, this allows the both of communication to remain restricted to proximal locations while still enabling models learned in farther clusters to be acquired.

\subsection{Centralized to Decentralized Spectrum}
The FedAvg is on the centralized end of the FL algorithms spectrum whereas GFL, D2DFL and CD2DFL are on the decentralized end. HFL, HD2DFL and HGFL avail a mix of centralized and decentralized operations. iCFL and iCD2DFL, despite being fully decentralized, mimic hierarchical behavior at the device level. The boundaries between centralized and decentralized operations FL algorithms thus remain fluid allowing present and future algorithms to benefit from both types of interactions.

\begin{algorithm}[H]
\DontPrintSemicolon
    \KwInput{Model parameters $\theta_k$ for each node, degree $n_k$}
    \KwOutput{D2D Aggregated Models}
    Devices $i = 1...... \bm{C}$ chosen as Cluster Head (CH) of cluster $\mathcal{C}_i$\;
    \For{free nodes $k = 1...... \bm{N}-\bm{C}$}{
    Join $\mathcal{C}_i \ \forall \ i \in \bm{C}$ as a Cluster-Member (CM)\;}
    \For{Cluster Head $\mathcal{C}_i \forall \ i = 1 \dots \dots \bm{C}$ in parallel}
    {Send model to CM $\{p \ : \ p \in \mathcal{C}_i\}$  \;
     \For{Nodes $k\ =\ 1\ \dots\dots \bm{N}$ \textbf{in parallel}}{
     $\bm{\theta}^{t + \tau}_k \longleftarrow $ Model-Update ($\bm{\theta}_k^t$) \;
    $\bm{\theta}^{t+T}_k \longleftarrow $ D2D-Aggregate ($\bm{\theta}_k^{t + \tau}, n_k$) \;
    Share model with CH $\mathcal{C}_i$ with probability $d_k$ \;
    }
    Aggregate models received from CM $\{p \ : \ p \in \mathcal{C}_i\}$ \;
    $\bm{\theta}^{t+\beta}_{\mathcal{C}_i} = \bm{\theta}^{t+ T}_{{\mathcal{C}_i}}
        +  \sum\limits _{j \in \mathcal{C}_i} \eta_j \bm{\theta}_j^{t +T} $ \;
    
    $\bm{\theta}_{\mathcal{C}_i}^{t+1} \longleftarrow Gossip \ Exchange \ (\bm{\theta}^{t+1}_{\mathcal{C}_i})$
    }
    \SetKwFunction{FUpdate}{Model-Update}
    \SetKwFunction{DDAgg}{D2D Aggregate}
    \SetKwFunction{GossipEx}{Gossip Aggregate}
      \SetKwProg{Fn}{Function}{:}{}
      \Fn{\FUpdate{$\bm{\theta}_s^t$}}{
        \For{batch $b$ of size $\mathcal{M} \in \mathcal{D}_s $}{
        $\xi_b = (x_b, y_b)$ is the data sample pairs in $b$\;
        $\bm{\theta_{s}^{t+\tau}} = \bm{\theta}_{s}^t - \alpha_{s}^t \nabla \mathcal{L}_s(\xi_b; \bm{\theta}_{s}^t)$ \;
        }
        \KwRet $\bm{\theta}_k^{t+\tau}$\;
        }
    
    \Fn{\DDAgg{$\bm{\theta}_l^t,\ n_l$}}{
        Exchange models with 1-hop neighbors $n_l$ with probability $d_l$ \;
        Perform weighted aggregation : $\bm{\theta}^{t+\delta}_l = \bm{\theta}^{t + \tau}_l
            +  \sum\limits _{j \in n_l} \eta_j \bm{\theta}_j^{t +\tau} $ \;
        \KwRet $\bm{\theta}^{t+\delta}_l$
            }              
    \Fn{\GossipEx{$\bm{\theta}_{\mathcal{C}}^t$}}{
    \For{$g =1 \dots \dots \zeta$ gossip rounds}{
    Perform aggregation handshake with a randomly selected $\mathcal{C}_j$ \; 
    Exchange and aggregate models between the pair \;
    $\bm{\theta}^{t+1}_i = \bm{\theta}^{t+ \tau}_i
        +  \eta_j \bm{\theta}_j^{t + \tau} $ \;}
    \KwRet $\bm{\theta}^{t+1}_i$
    }
\caption{Inter-Cluster Device-to-Device Federated Learning (iCD2DFL)}\label{alg:iCD2DFL}
\end{algorithm}

\begin{figure*}[t]
    \centering
    \includegraphics[width = \textwidth, height = 8cm]{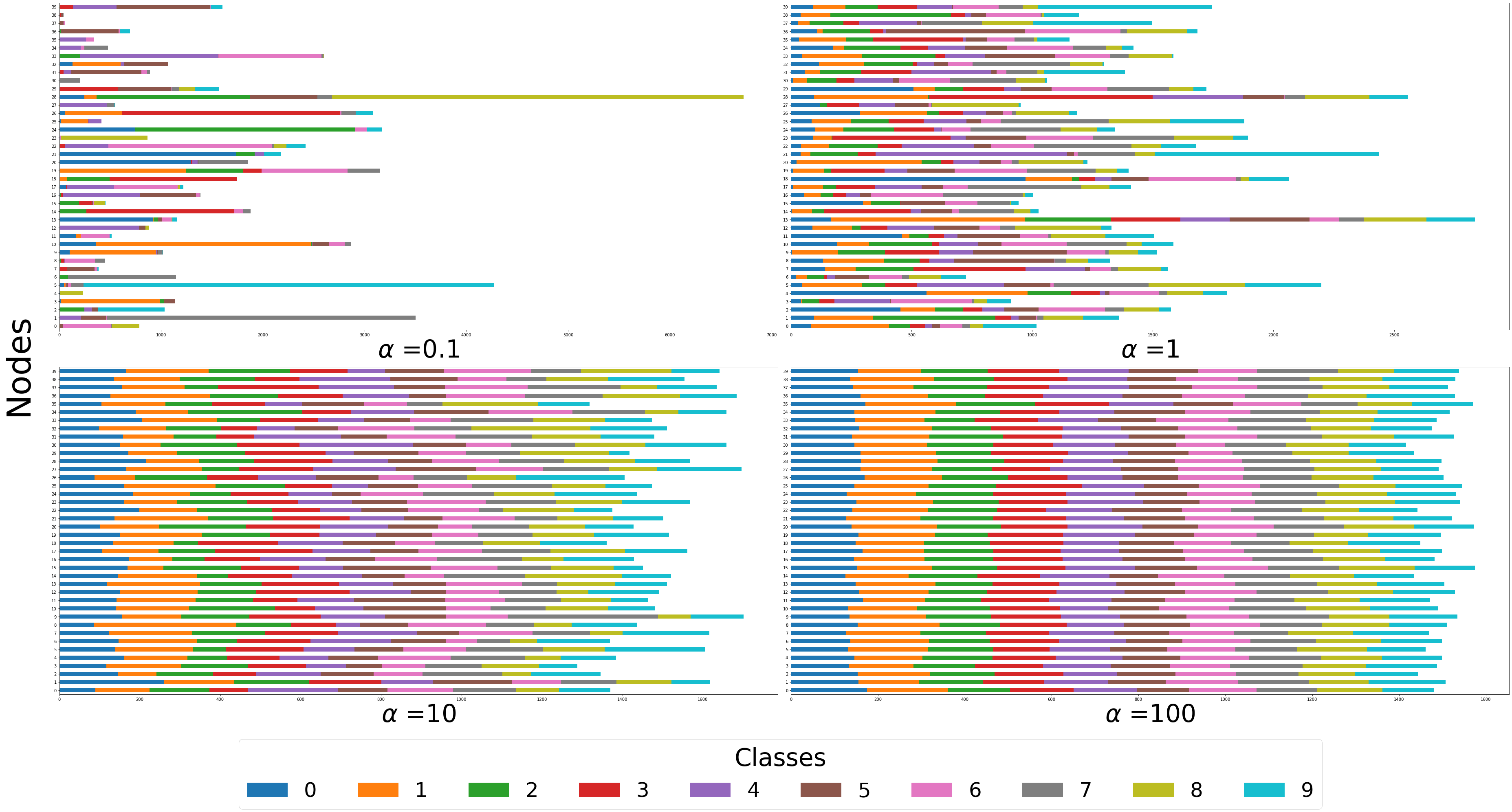}
  \caption{Synthetic data distributions for MNIST dataset across 40 nodes derived from Dirichlet Distribution for different $\alpha$ values. 
    }
    \label{fig:datadist}
\end{figure*}

\section{Experiments and Evaluation} \label{sec:expeval}
A uniform evaluation strategy was devised to conduct an objective assessment of the various FL algorithms. The experiments performed in this work use the MNIST \cite{deng2012mnist} and FashionMNIST datasets. \cite{xiao2017fashion} 
A non-IID data distribution remains of primary interest in this work. The experimentation included subjecting the FL algorithms to ideal and noisy communication, probabilities of participation, data distribution and aggregation frequency. Additionally, these algorithms were also subjected to Few-Shot Learning details of which have been shared subsequently. 

\paragraph{Non-IID Distribution} The non-IID data partitions are generated using the Dirichlet Distribution \cite{li2021model} parametrized by its concentration parameter $\alpha$. 
The value of $\alpha$ controls the degree of non-IID sampling spread across the clients with lower values resulting in higher imbalance. Lower values of $\alpha$ result in more skewed datasets as shown in Fig. \ref{fig:datadist}. A value of $\alpha$ between 1 and 10 results in a typically non-IID data distribution whereas lower values of $\alpha < 1$ result in extreme non-IID distributions. On the other end, values of $\alpha > 100$ result in increasingly IID distributions. The evaluation in this work uses $\alpha = 0.1$ and $\alpha = 1.0$ for generating highly non-IID distributions. 
\paragraph{Extremely Skewed non-IID Distribution} This work also employs an extremely skewed 2-class and 3-class non-IID cases 
where each node is trained using data samples from two and three classes only respectively. Each node is randomly assigned the determined number of classes and the individual class data is grouped into \textit{shards} of 50 images each. The nodes are then assigned a randomly chosen number of shards sampled from their respective classes.

\paragraph{Neural Network and Development Environment}
The DNNs used for evaluations based on MNIST and Fashion MNIST datasets are comprised of two 2D–Convolutional blocks followed by a Dropout and two linear layers. 
The modes are instantiated with the same weights while training is conducted using the SGD optimizer with a 
learning rate $\eta = 0.01$. The entire framework has been developed in Python whereas the learning algorithms were implemented using PyTorch. The neural network trained for MNIST and Fashion MNIST contains approximately 2 million parameters 
oth neural networks use Rectified Linear Unit (ReLU) activations with all layers except the final layer which 
uses Log-Softmax activation to generate the 
class probabilities. 

\paragraph{Training Regime}
The training environment assumes a network of 40 nodes, all initiated with similar weights. Our framework allows for each node to control the number of local updates as well as the learning rate and other hyperparameters. However, for this work,  only variation in number of training epochs has been conducted across the devices. 

\paragraph{Operating Characteristics}
The network topology generated for the experiments consists of $\bm{N} = 40$ nodes that are divided into $\mathcal{C} = 7$ clusters. In order to ensure \textit{reachability}, $\gamma = 0.95$ and $\upsilon = 0.1$ have been used. The former ensures that each device has a direct link to ~95\% of the devices from the same cluster. Additionally, 10\% of each devices' neighbors lie outside the cluster. This ensures that the groups have sufficiently dense intra-device connections as well as having links with nodes outside the current group. This resulting network graph remains \textit{reachable}.
The experiments conducted during the course of this work encompass two different participation scenarios:
\begin{enumerate*}[label = (\alph*)]
    \item Upstream participation with probability $p_k$ and
    \item Neighborhood participation with probability $d_k$
\end{enumerate*}
The participation probabilities of $p_k = \{0.9, 0.6\}$ and $d_k = \{0.9, 0.6\}$ have been used during the course of this work. The two conditions have also been tested jointly to mimic scenarios of frequent stragglers and communication limitations.The channel conditions for both D2D and D2E communication have been subjected to zero-mean Gaussian noise $\mathcal{N}(\mu = 0, \sigma^2)$\footnote{$\bm{N}(0, \sigma^2)$ noise instead of more detailed fading models and frequency-selective channel impairments has been considered since the primary aim remains to gauge the impact of noisy updates being used for aggregation.} with $\sigma^2 = 0.01, 0.0025, 0.0001$. 

\paragraph{Few-Shot Federated Learning} One-Shot FL \cite{guha2019one} suggests training local models to completion before sharing with the global server. The resulting mechanism enables significant reduction in communication frequency. Extending the idea, this work subjects the FL algorithms to Few-Shot FL. The nodes perform significantly more local updates before sharing their models. The scheme works with significantly reduced aggregation rounds and communication at both device and edge level. However, it also results in higher client drift.

\paragraph{Asynchronous Operation}
FL envisions autonomy of operation at the device level. By extension, this implies that not all of the devices perform same number of update operations during the training phase. This may be caused by device being active (or inactive) or a straggler. This work assumes that the devices are required to work with a deadline after which they are required to share their respective models for aggregation. This deadline-based asynchronous aggregation behavior is replicated for the FL algorithms by allowing each device to undergo a range-limited random number of training epochs. 


\begin{figure*}[t]
\centering
\def\twidth{0.90}
\subfloat[$\alpha = 1.0$ ]{%
\includegraphics[scale =0.22]{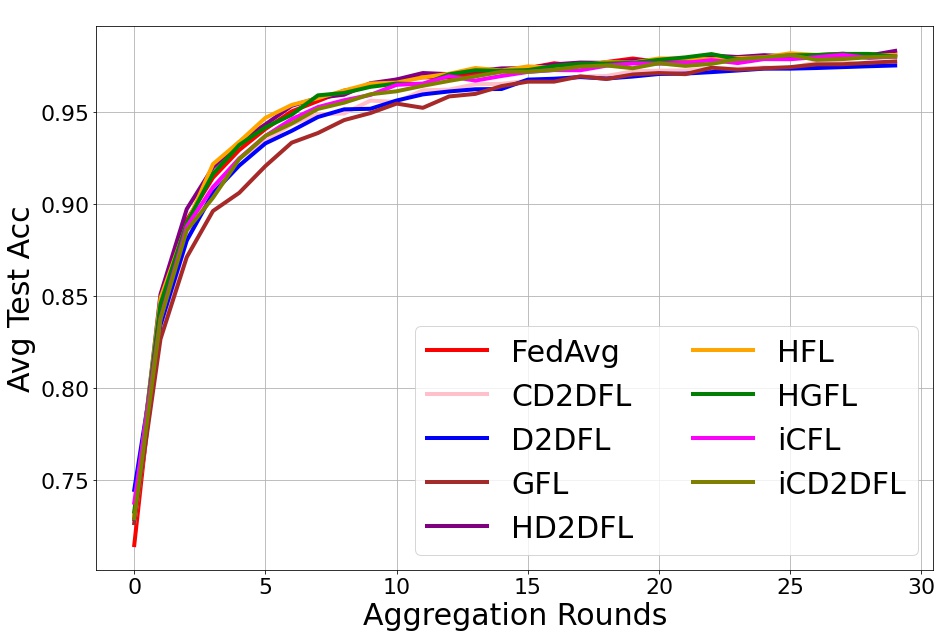}%
}
\subfloat[$\alpha = 0.1$]{%
\includegraphics[scale =0.22]{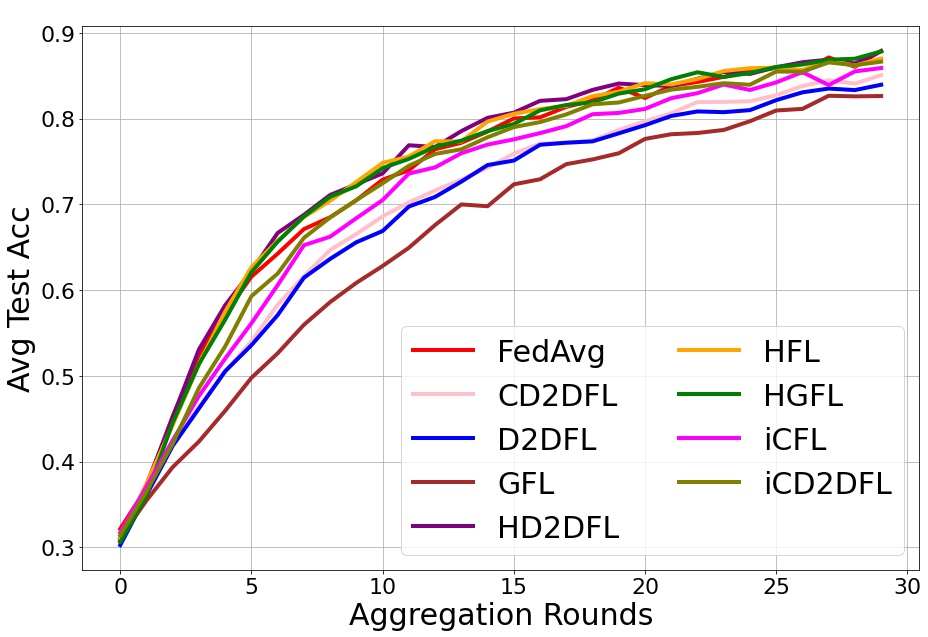}%
}\hfill
\subfloat[$\alpha = 1.0$ ]{%
\includegraphics[scale = 0.22]{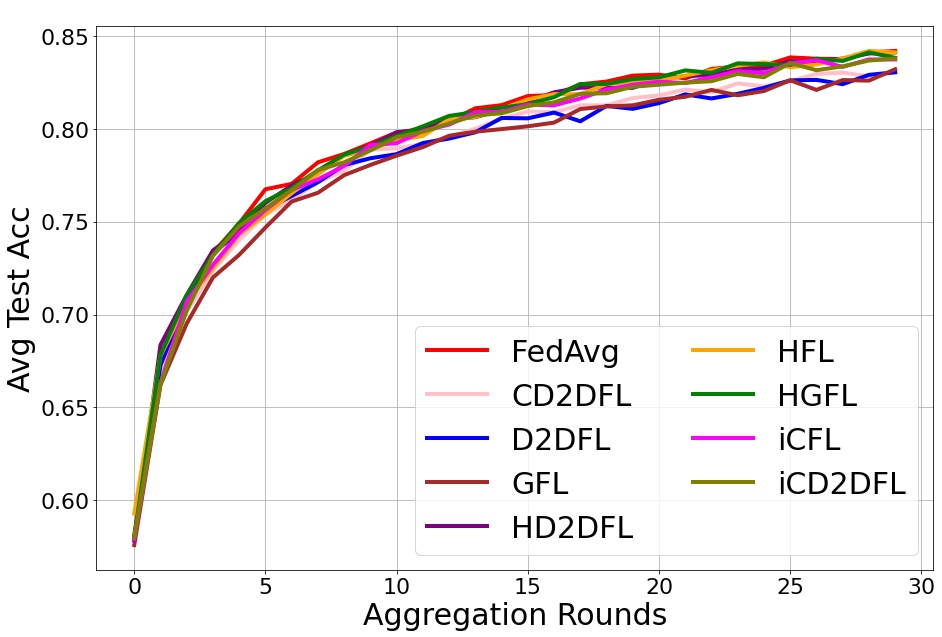}%
}
\subfloat[$\alpha = 0.1$]{%
\includegraphics[scale =0.22]{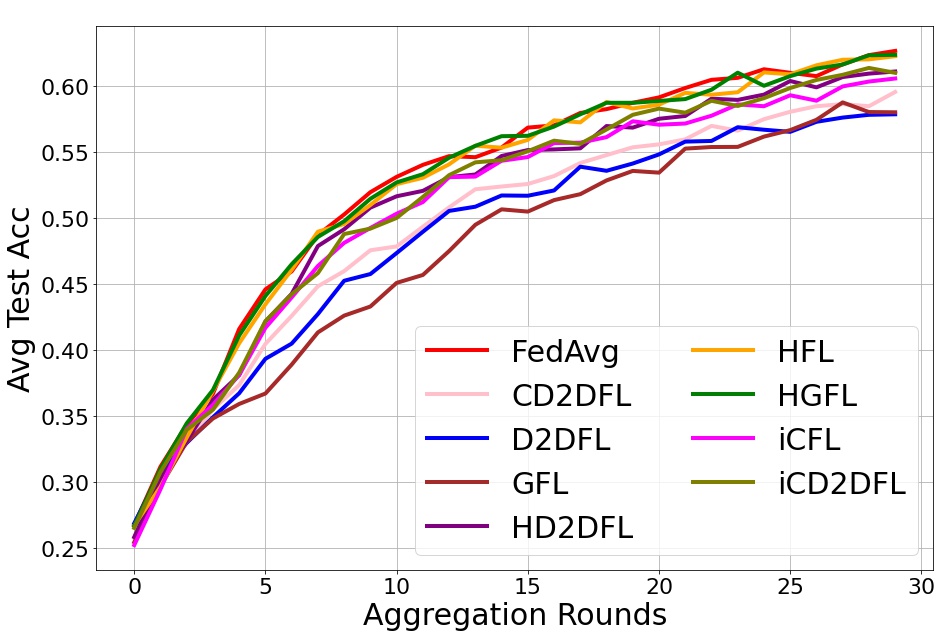}%
} \hfill
\caption{Average Test Accuracy for (a)-(b) MNIST and (c)-(d) FashionMNIST for maximal participation $p_k, d_k = 0.9$ with  asynchronous communication for Non-IID distributions with Dirichlet parameter $\alpha = 10 $ and $\alpha =0.1$}
\label{fig:maxpart}
\end{figure*}
\section{Performance Analysis} \label{sec:analysis}

\begin{figure*}[t]
\centering
\def\twidth{0.90}
\subfloat[$\alpha = 1.0$]{%
\includegraphics[scale = 0.2]{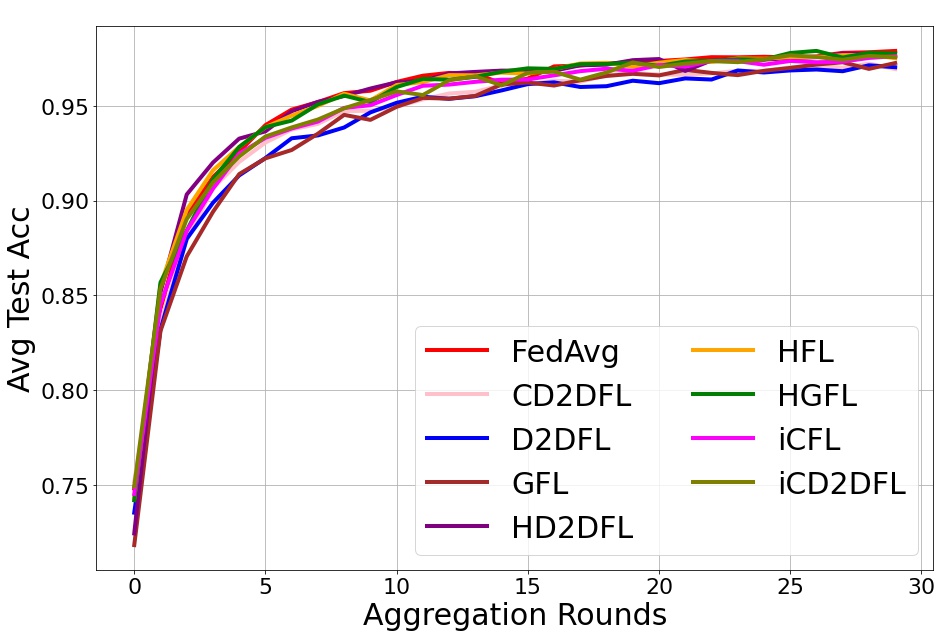}%
}
\subfloat[$\alpha = 0.1$]{%
\includegraphics[scale =0.2]{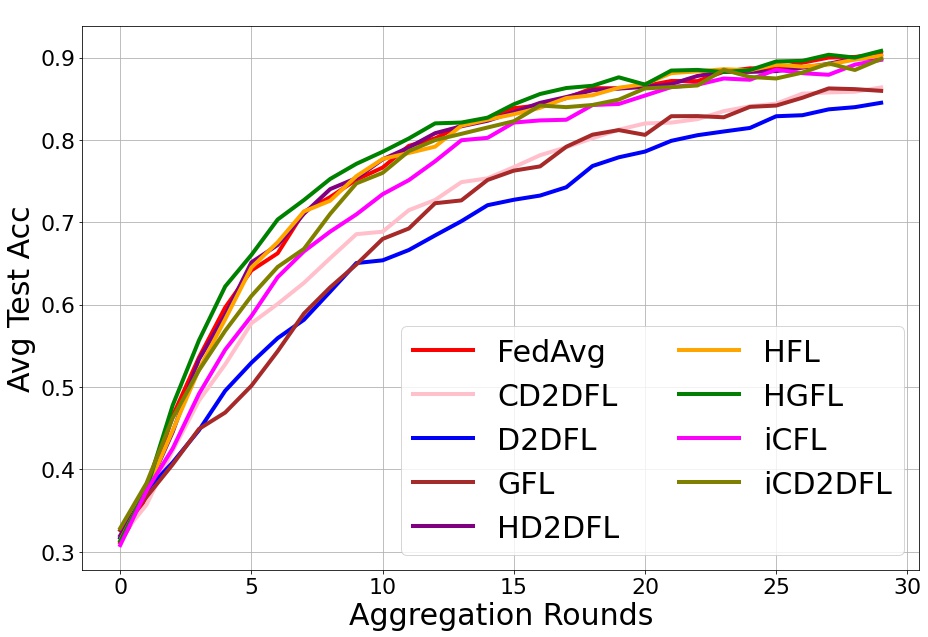}%
}\hfill
\subfloat[$\alpha = 1.0$]{%
\includegraphics[scale = 0.2]{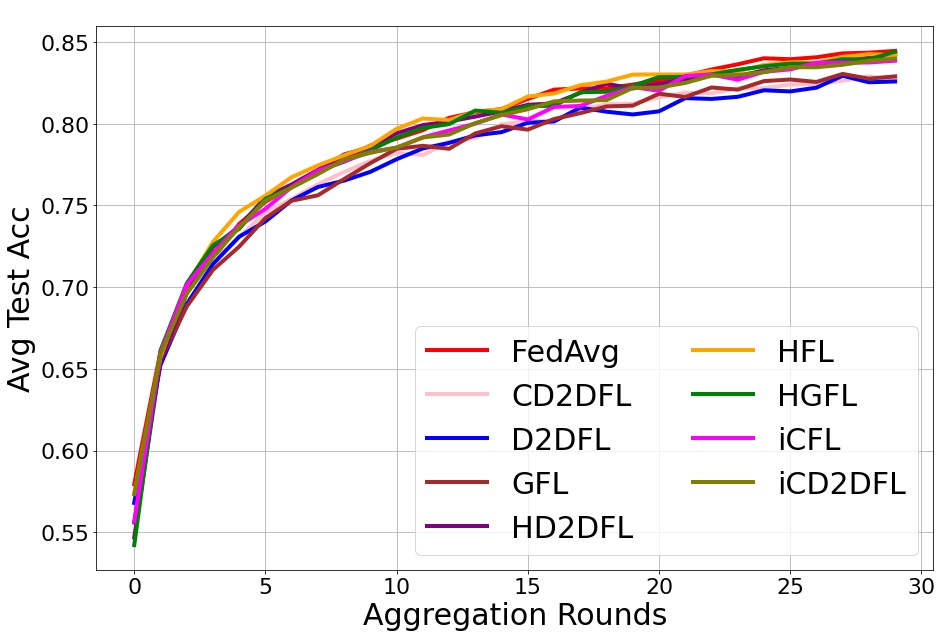}%
}
\subfloat[$\alpha = 0.1$]{%
\includegraphics[scale =0.2]{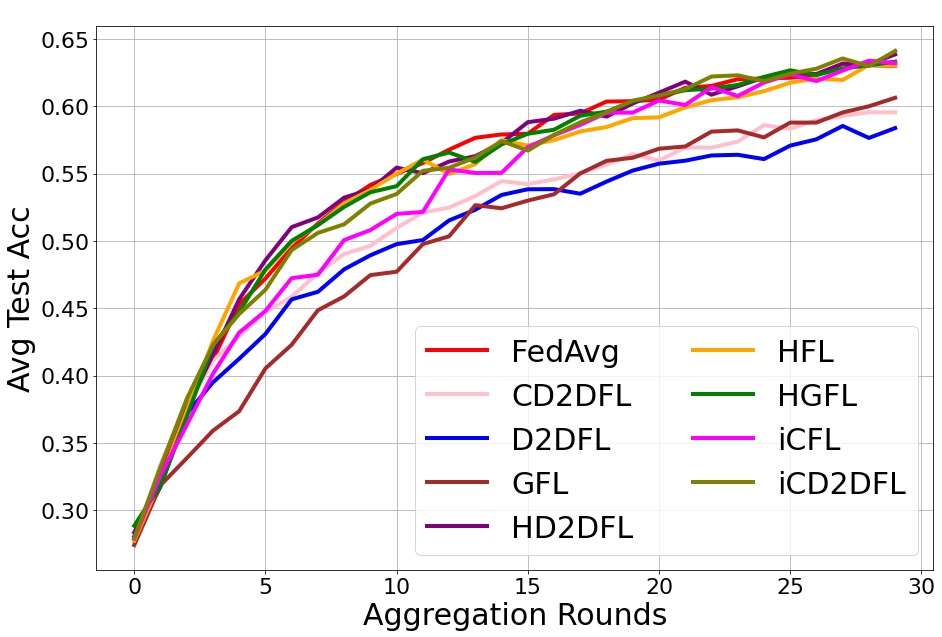}%
} \hfill
\caption{Average Test Accuracy for (a)-(b) MNIST and (c)-(d) FashionMNIST for limited participation $p_k, d_k = 0.6$ with  asynchronous communication for Non-IID distributions with Dirichlet parameter $\alpha = 1.0 $ and $\alpha =0.1$}
\label{fig:redpart}
\end{figure*}

\begin{figure*}[!ht]
\centering
\def\twidth{0.90}
\subfloat[2-class non-IID]{%
\includegraphics[scale = 0.2]{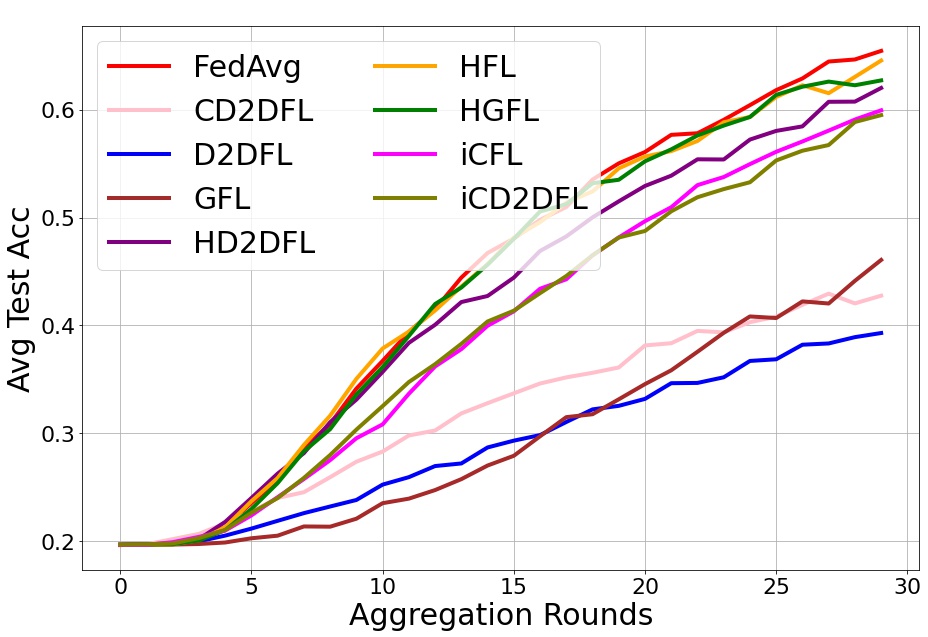}%
}
\subfloat[3-class non-IID]{%
\includegraphics[scale =0.2]{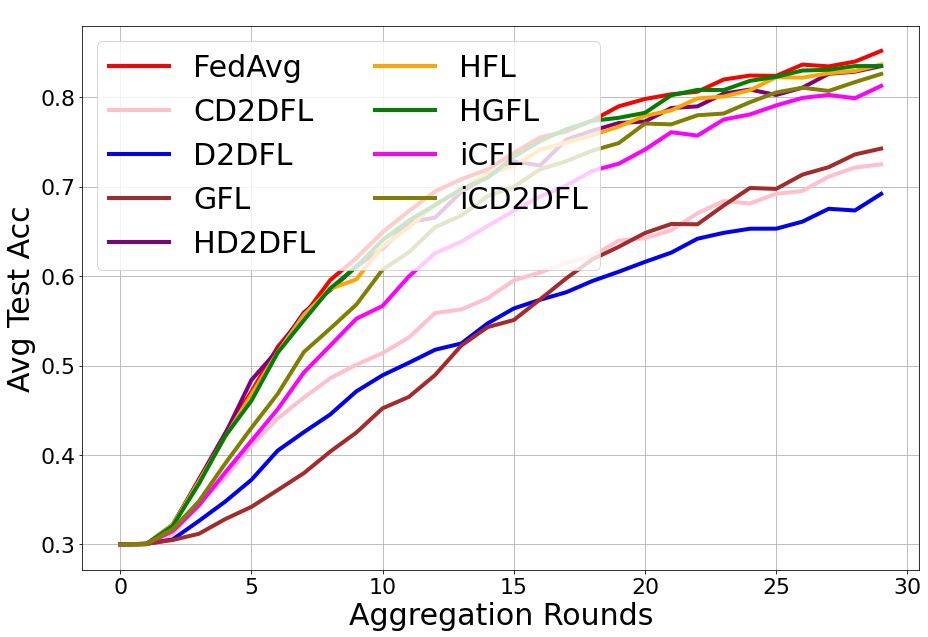}%
}\hfill
\subfloat[2-class non-IID]{%
\includegraphics[scale = 0.2]{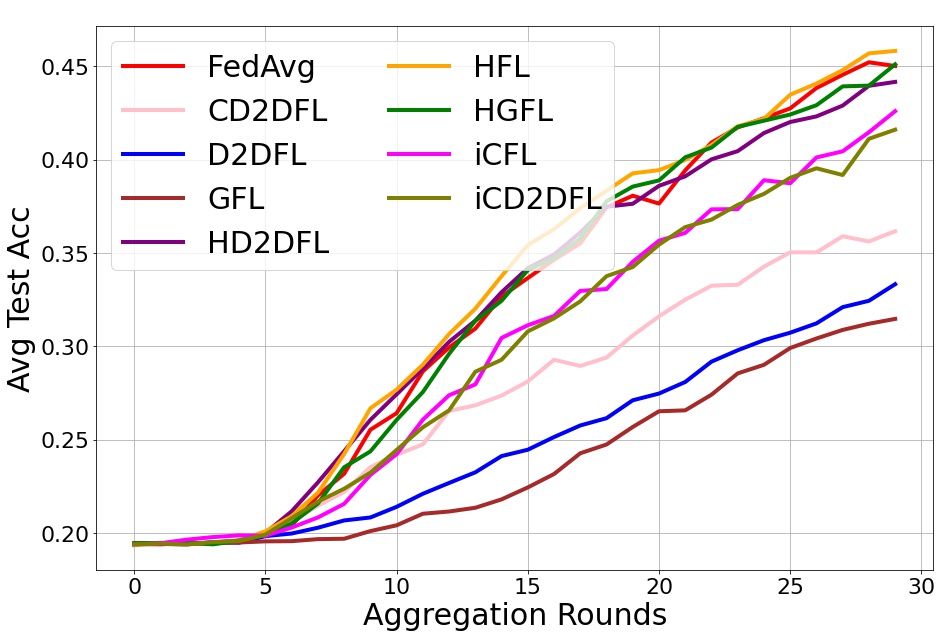}%
}
\subfloat[3-class non-IID]{%
\includegraphics[scale =0.2]{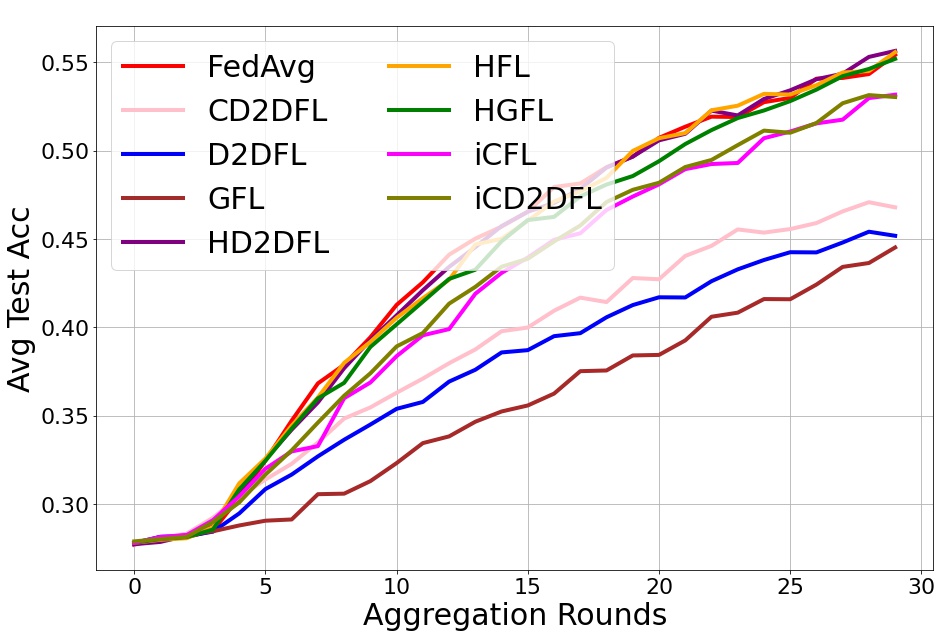}%
} \hfill
\caption{Average Test Accuracy for (a)-(b) MNIST and (c)-(d) FashionMNIST for limited participation $p_k, d_k = 0.6$ with  for extremely skewed (2-class and 3-class) Non-IID distributions. }
\label{fig:redpart_skew}
\end{figure*}

The following section encapsulates the findings for each of the mentioned simulation conditions. The final test accuracies of each algorithm, averaged over three runs and over all the nodes have been reported in Figures~ \ref{fig:maxpart}-\ref{fig:fewshot}. Since, one of the aims of decentralized FL is to reduce the communication cost, we also present the number of messages for each link type for each algorithm in Fig.~\ref{fig:acc-comm}. 


\subsection{FL with Maximal Participation and Ideal Communication}
The ideal scenario assumes noise-free communication and a $90\%$ device participation probability i.e. $p_k, d_k > 0.9$. As expected, greater non-IID distribution across device adversely affects the convergence rate and performance. The impact is reflected across all algorithms. However, results for FashionMNIST depict more loss in performance. This indicates a strong correlation between the performance degradation jointly due to a DNN's ability to learn and greater non-IID levels.  The results in Fig. \ref{fig:maxpart} indicate that decentralized FL performs on par with their more centralized counterparts. The results also show that as distributions become more non-IID, GFL, D2DFL and CD2DFL lag in learning performance when compared to the other algorithms. However, GFL also incurs the least communication volume among all the considered FL schemes since it only uses the least costly communication link as shown in  Fig.~\ref{fig:acc-comm}. \\
The decentralized algorithms also indicate a slightly slower convergence rate exacerbated as the training progresses. However, clustered operations, particularly iCD2DFL and iCFL indicate better convergence and consistently perform better than D2DFL and GFL. This indicates that inter-cluster cooperation orchestrated by these algorithms helps in the learning performance and may help in arresting overtraining even in the absence of global information. This is particularly evident from iCFL where D2D interaction at the device level is absent.

\subsection{FL with Limited Device Participation}
The results in Fig. \ref{fig:redpart} present the results with reduced device participation. Less number of devices joining an aggregation round may be caused due to stragglers or busy devices. All the algorithms suffer from convergence issues as the device participation is decreased. This implies that data distribution and device participation have confounding effects on the learning performance of these algorithms. The difference in performance of the D2DFL, GFL and CD2DFL becomes more pronounced than the rest as the training proceeds. The performance suffers clearly from overtraining compounded by lower number of devices as well highly non-IID data. The convergence rate of these three algorithms slows more evidently than the others as the training proceeds. On the other hand, inter-cluster operations allow iCFL and iCD2DF to perform better than the others even with 60\% device participation.

The results from Fig. \ref{fig:redpart} are reinforced by testing the FL algorithms under limited device participation and extremely skewed data distribution as shown in Fig. \ref{fig:redpart_skew}. When subjected to 2-class and 3-class non-IID data distributions, the localized operation in D2DFL, GFL and CD2DFL suffers more than the others. Lesser number of classes result in almost 20\% performance difference between these algorithms and the rest as indicated in Fig. \ref{fig:redpart_skew}(a) and (c). The performance of iCFL and iCD2DFL, however, remains remarkably robust even in the presence of these extremely adverse conditions. Both algorithms, supported by inter-CH operation, perform within 5\% of the centralized and hierarchical algorithms. The convergence rate of these algorithms retains its trajectory while other decentralized algorithms show greater divergence as the training proceeds. It may therefore be inferred that gossip operations by the CHs partially offset the degradation caused by the lesser device participation and extremely non-IID data and acts akin to DropOut operations in a DNN.
Table. \ref{table:results} provides an accuracy-based performance overview of all the algorithms both with maximal and limited participation conditions.

\begin{table}[H]
\centering
\def\twidth{0.90}
\begin{tabular}{cllll}
\toprule
\multirow{2}{*}{\textbf{Algorithm}} & \multicolumn{2}{c}{$p_k = 0.6, d_k = 0.6 $} & \multicolumn{2}{c}{$ p_k = 0.9, d_k = 0.9 $} \\ \cmidrule{2-5}
                          & $\alpha = 1.0$   & $\alpha =0.1$  & $\alpha = 1.0$ & $\alpha =0.1$ \\ \midrule
FedAvg                        & 0.9794             & 0.8736       & 0.9786                & 0.8810      \\ \hline
D2DFL                      & 0.9758        & 0.8454        & 0.9752           & 0.8849      \\
HFL                        & 0.9805*        & 0.9022*       & 0.9808        & 0.9257*      \\
HD2DFL                     & 0.9798*        & 0.8980       & 0.9831*           & 0.9189      \\
GFL                        & 0.8744        & 0.3824       & 0.9753         & 0.8398      \\
HGFL                       & 0.9801*      & 0.9015*  & 0.9800 & 0.9241      \\
iCFL                      & 0.9762       & 0.8972       & 0.9791   & 0.9142      \\
CD2DFL                    & 0.9694       & 0.8651       &  0.9730  & 0.8855 \\
iCD2DFL                   & 0.9803*      & 0.8983       & 0.9800       & 0.9162      \\ \bottomrule
\end{tabular}
\caption{Accuracy comparison of FL algorithms for two sets of $p_k, d_k$ values and different values of the Dirichlet parameter $\alpha$ (lower values indicate increased non-IID distribution). Highest accuracies within $\Delta = 0.001$ have been marked with asterisk (*)}
\label{table:results}
\end{table}

\subsection{FL with Noisy Communication}
\begin{figure*}[t]
\centering
\def\twidth{0.90}
\subfloat[$\sigma^2 = 0.0001$]{%
\includegraphics[scale = 0.15]{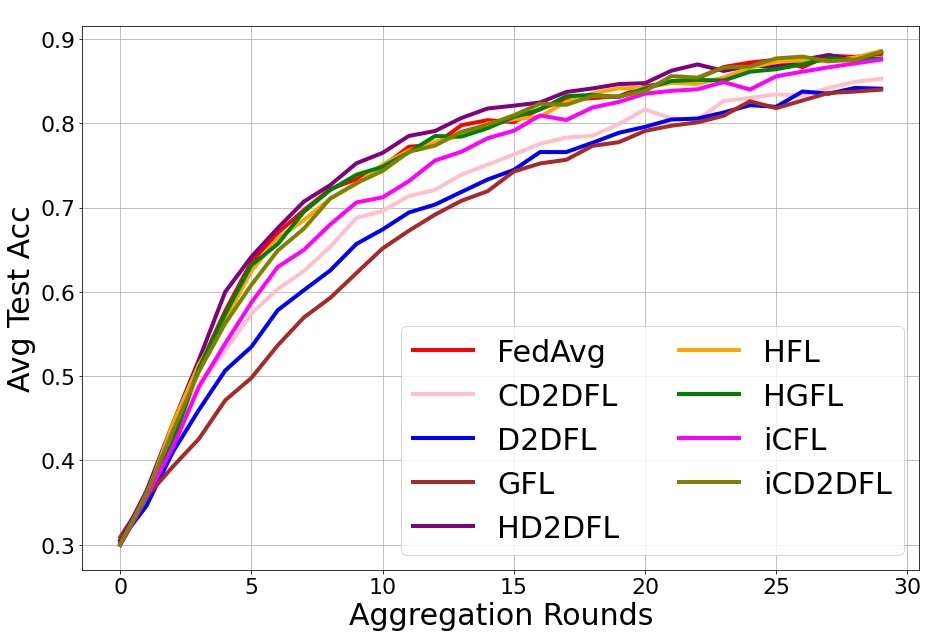}%
}
\subfloat[$\sigma^2 = 0.0025$]{%
\includegraphics[scale =0.15]{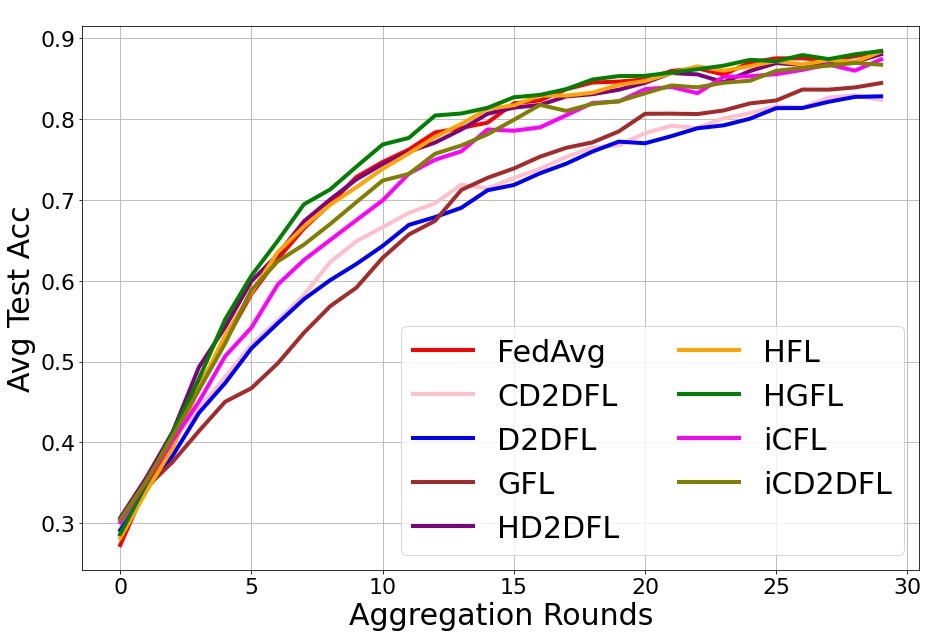}%
}
\subfloat[$\sigma^2 = 0.01$]{%
\includegraphics[scale =0.15]{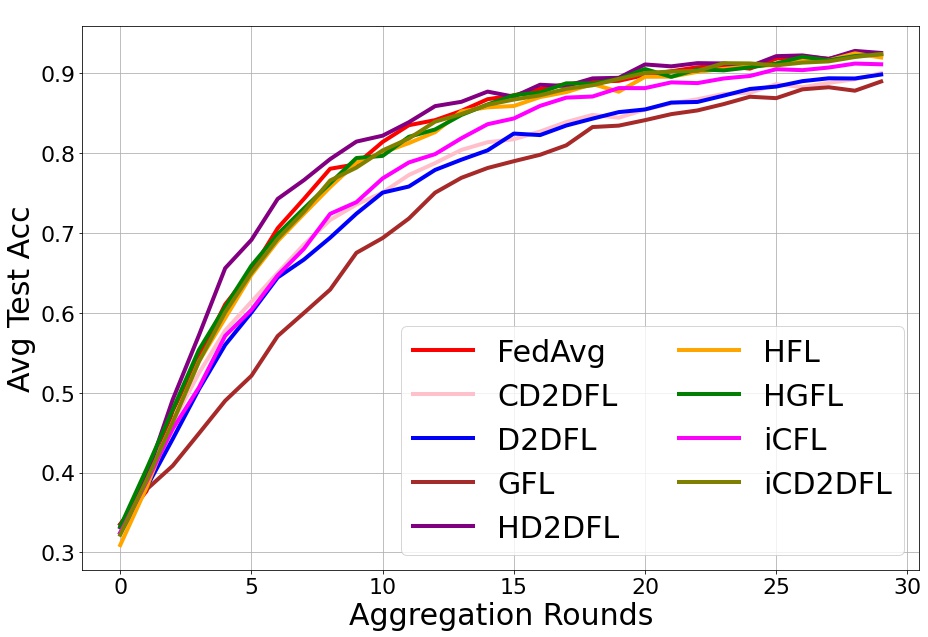}%
}
\hfill
\subfloat[$\sigma^2 = 0.0001$]{%
\includegraphics[scale = 0.15]{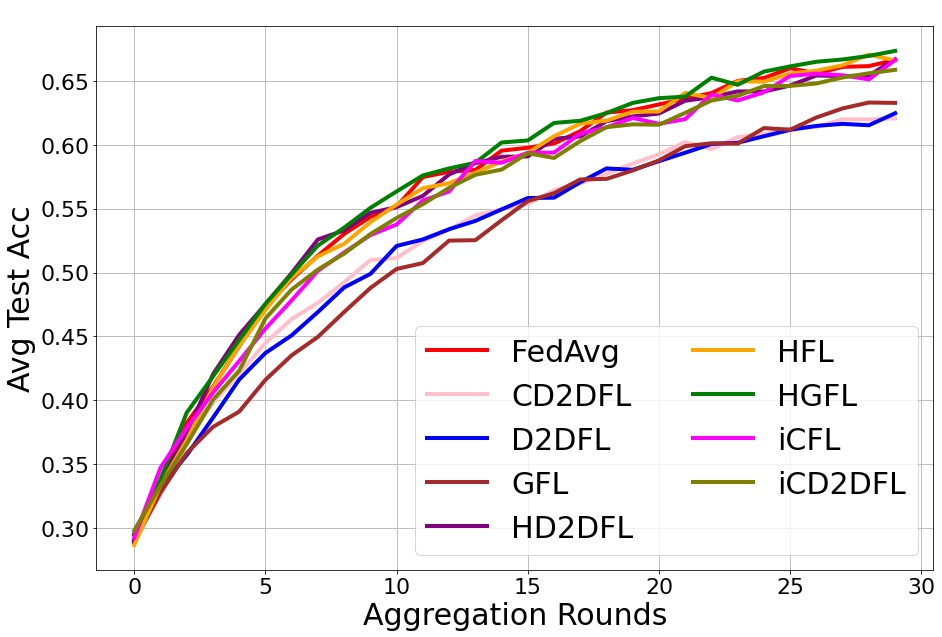}%
}
\subfloat[$\sigma^2 = 0.0025$]{%
\includegraphics[scale =0.15]{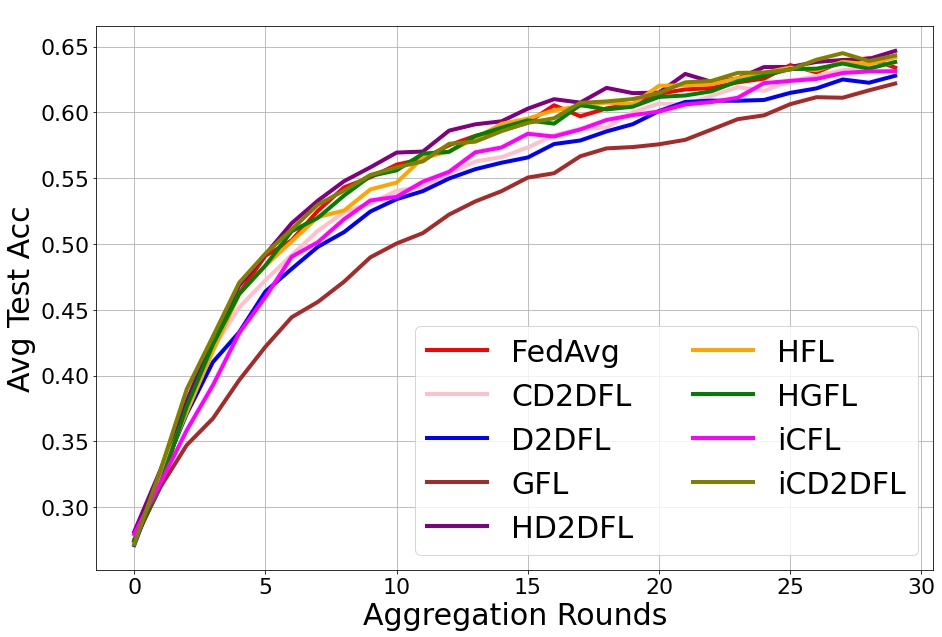}%
}
\subfloat[$\sigma^2 = 0.01$]{%
\includegraphics[scale =0.15]{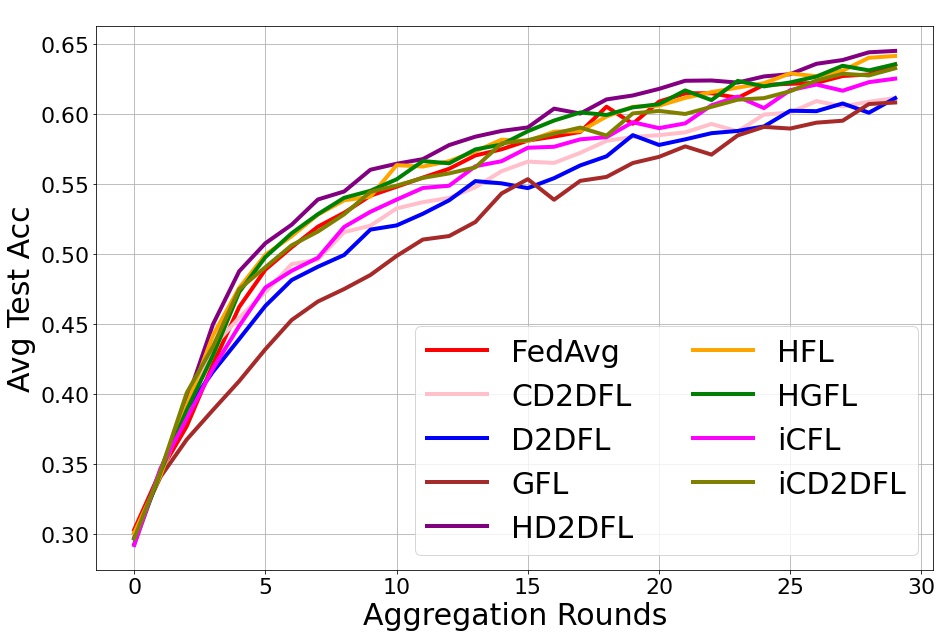}%
}
\hfill
\caption{Average Test Accuracy for (a)-(c) MNIST and (d)-(f) FashionMNIST for $\bm{N}(0, \sigma^2)$ noisy communication with $p_k, d_k = 0.9$ and Dirichlet parameter $\alpha = 0.1$.}
\label{fig:noisy}
\end{figure*}

The largest application of Federated Learning is envisioned in wireless spectrum. However, wireless media is subject, among other challenges, to a significant presence of noise. With FL requiring to exchange hundreds of thousands of parameters, presence of noise may cause the convergence to slow down considerably. The FL algorithms during the current research were subjected to the presence of Gaussian Noise$~\mathcal{N}(0, \sigma^2)$ 
both at the device communication level and during D2E and E2C communication. From Fig. \ref{fig:noisy} 
it can be observed that the addition of noise at the aggregation stage mildly slows down convergence. HGFL progressively shows reduction in performance as $\sigma^2$ is increased. However, clustering operation in iCFL and iCD2DFL allow them to closely follow the performance by HD2DFL and HFL. In contrast, GFL shows reduced convergence as compared to D2DFL. However, as training progresses, D2DFL and CD2DFL both also indicates a plateauing convergence rate. 

\begin{figure*}[t]
\centering
\def\twidth{0.90}
\subfloat[$\alpha = 0.1$ MNIST]{%
\includegraphics[scale = 0.22]{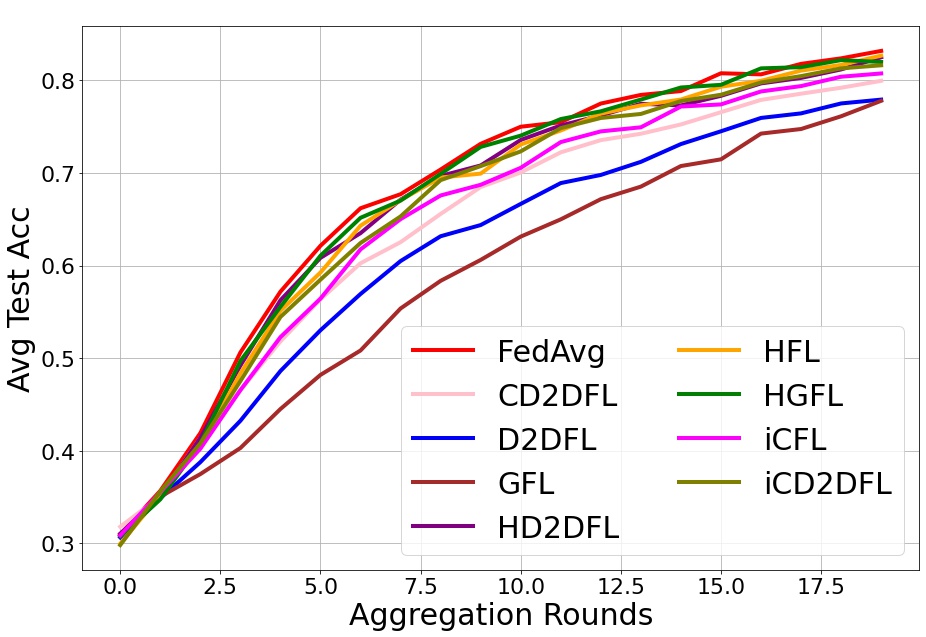}%
}
\subfloat[$\alpha = 0.1$ FashionMNIST]{%
\includegraphics[scale =0.22]{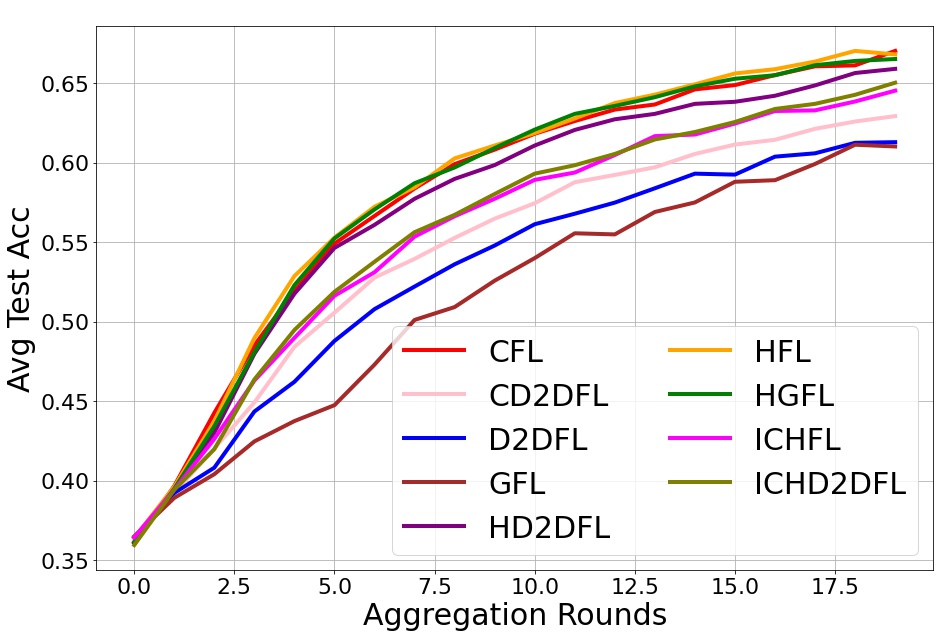}%
}\hfill
\subfloat[$\alpha = 0.1$ MNSIT]{%
\includegraphics[scale = 0.22]{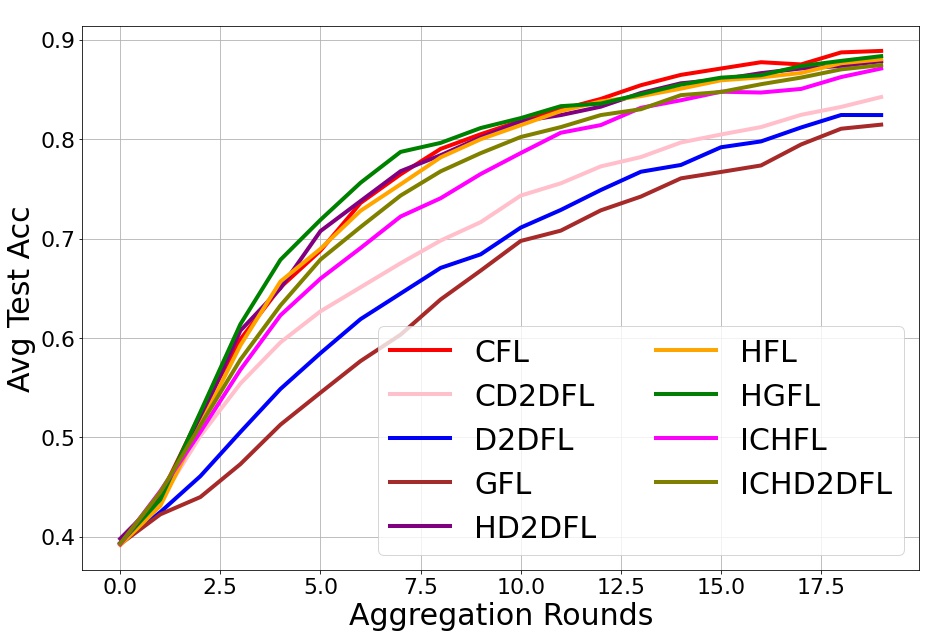}%
}
\subfloat[$\alpha = 0.1$ FashionMNIST]{%
\includegraphics[scale =0.22]{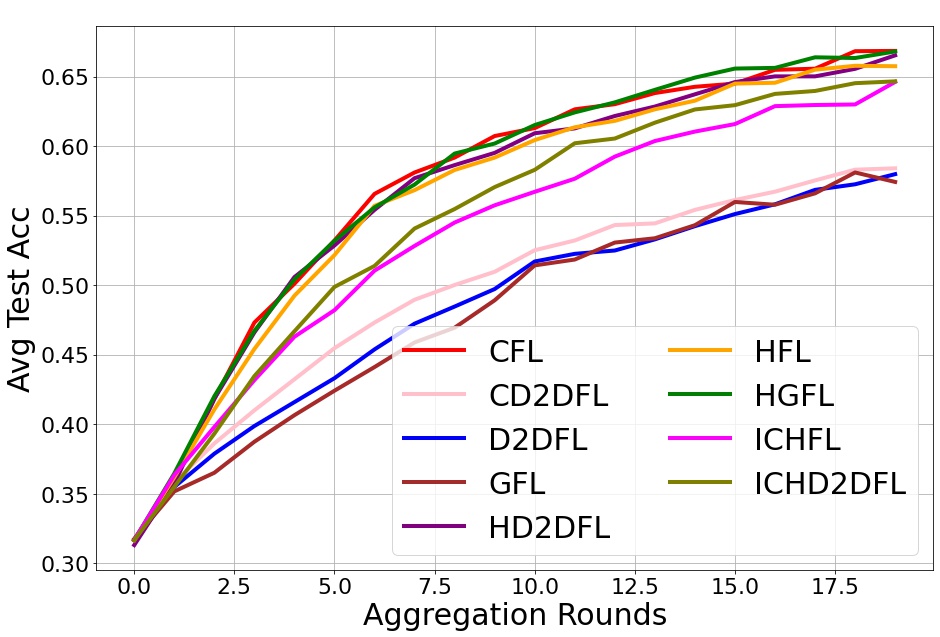}%
} \hfill
\caption{Average Test Accuracy for Few-Shot Learning for MNIST and FashionMNIST with (a)-(b) $p_k, d_k = 0.9$ and (c)-(d) $p_k, d_k = 0.6$ with $r =20$ aggregation rounds and epochs in range $[15, 20]$ with  asynchronous communication for Non-IID distributions with Dirichlet parameter $\alpha =0.1$}
\label{fig:fewshot}
\end{figure*}

\subsection{Few Shot Learning}
Few-Shot FL offers one way of reducing the communication cost when applied properly. The devices undergo multiple local updates before sharing their models for aggregation. Fig. \ref{fig:fewshot} shows the result of Few-Shot learning applied under maximal and limited device participation scenario with $p_k, d_k = 0.9$ in Fig. \ref{fig:fewshot}(a)-(b) and $p_k, d_k = 0.6$ in Fig. \ref{fig:fewshot} (c)-(d). The additional difference in performance can be attributed to greater non-IIDness in figures (c) and (d). The impact of client drift on all algorithms causing the learning to plateau earlier than regular learning regime is evident when its results are compared with the ones shown in Fig. \ref{fig:maxpart}-\ref{fig:redpart}. Overall, the decentralized variants are impacted worse than the centralized and hierarchical algorithms. Additionally, the performance of D2DFL, GFL and CD2DFL lags further than the rest as training progresses. The performance of iCFL and CD2DFL while showing relatively bigger drop than when subjected to normal training scheme, still achieves an accuracy within 5\% of the centralized algorithms. The overall performance suggests scenarios where Few-Shot learning may be considered practicable when communication costs become prohibitive especially or shared updates are infrequent.

\begin{figure}[t]
    \centering
    \includegraphics[scale = 0.5]{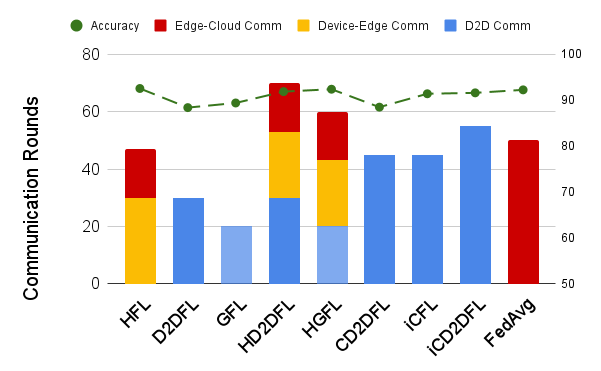}
    \caption{Average FL Algorithm Accuracy and per-Node Communication Volume for asynchronous aggregation for 30 rounds with the MNIST dataset.}
    \label{fig:acc-comm}
\end{figure}

\subsection{Communication Cost and Volume}
When dealing with different FL algorithms, it is essential that the difference between links employed during the operation is kept in purview. The work in \cite{7343681} suggest that D2D links like LTE-Direct, WiFi Direct or DSRC are more energy efficient than Device-Edge cellular links like LTE. Furthermore, the same research also indicates that the mentioned D2D links offer reduced latency when compared with cellular communication. E2C links must be be preceded by D2E links for any communication meant for the cloud. It may therefore be inferred that the energy cost and latency of uploading to the cloud may well be more than both D2D and D2E communications. It is with this understanding that Fig.\ref{fig:acc-comm} considers D2D links as least costly and E2C links as most expensive both in terms of energy and latency.

Fig-\ref{fig:acc-comm} shows the communication volume required by different FL algorithms. This is calculated by identifying the number of aggregation rounds required by each of the FL algorithms and are categorized according to the target level (device, edge or cloud). The centralized variants employ more expensive cellular communication for D2E communication. The additional load on network backbone in form of E2C links is bound to significantly increase the cost for FL albeit its improved performance. Fig.\ref{fig:acc-comm} shows the communication volume in terms of communication rounds for various FL algorithms with the accuracy achieved by the respective algorithms under non-IID data settings. The highest performing algorithms use the most expensive forms of communication. On the contrary, decentralized FL variants indicate slower convergence but also incur less communication cost.  

Comparably, decentralized scheme with inter-cluster aggregations show appreciable performance, falling slightly short of the accuracy of HGFL and HD2DFL.

\section{Conclusion}\label{sec:conclusion}
This work presents a comprehensive comparison of various FL algorithms that have so far been viewed only from the perspective of Centralized FL. The FLAGS framework developed for this purpose enables configuring multiple FL architectures expeditiously. The contrasting highlight of this work is a detailed comparison of major FL algorithms under some of the most dominant challenges in this domain. frameworks.
The analysis conducted here indicates that decentralized FL performs comparatively well despite no or minimal upstream communication. Even though noisy communication and irregular participation negatively impacts decentralized FL performance, such modes are still capable of achieving acceptable performance thresholds. However, extremely skewed data distributions degrade fully decentralized FL considerably more than centralized ones. The results indicate that FL modes may be used interchangeably depending on the network conditions and the communication costs. As part of the future work, we also intend to investigate the performance of the FL algorithms for feature skewed non-IID distributions. Furthermore, based on the results presented here, the next milestone would be to formulate an adaptive mechanism based on the operating and environment characteristics where the nodes may be suggested to follow a particular FL algorithm for efficient operation.  

\bibliography{ref_VFL}

\begin{thebibliography}{10}
\expandafter\ifx\csname url\endcsname\relax
  \def\url#1{\texttt{#1}}\fi
\expandafter\ifx\csname urlprefix\endcsname\relax\def\urlprefix{URL }\fi
\expandafter\ifx\csname href\endcsname\relax
  \def\href#1#2{#2} \def\path#1{#1}\fi

\bibitem{FL_McMahan_Moore_Ramage_Hampson_Arcas_2017}
H.~B. McMahan, E.~Moore, D.~Ramage, S.~Hampson, B.~A.~y. Arcas,
  Communication-efficient learning of deep networks from decentralized
  dataArXiv: 1602.05629 (Feb 2017).

\bibitem{kairouz2019advances}
P.~Kairouz, H.~B. McMahan, B.~Avent, A.~Bellet, M.~Bennis, A.~N. Bhagoji,
  K.~Bonawitz, Z.~Charles, G.~Cormode, R.~Cummings, et~al., Advances and open
  problems in federated learning, arXiv:1912.04977 (2019).

\bibitem{ansari20175g}
R.~I. Ansari, C.~Chrysostomou, S.~A. Hassan, M.~Guizani, S.~Mumtaz,
  J.~Rodriguez, J.~J. Rodrigues, 5g d2d networks: Techniques, challenges, and
  future prospects, IEEE Systems Journal 12~(4) (2017) 3970--3984.

\bibitem{fullyFL_Lalitha_Javidi_Shekhar_Koushanfar}
A.~Lalitha, S.~Shekhar, T.~Javidi, F.~Koushanfar, Fully decentralized federated
  learning, in: Third workshop on Bayesian Deep Learning (NeurIPS), 2018.

\bibitem{gossip_FL_2019}
I.~Hegedüs, G.~Danner, M.~Jelasity, Gossip Learning as a Decentralized
  Alternative to Federated Learning, Vol. 11534 of Lecture Notes in Computer
  Science, Springer International Publishing, 2019, p. 74–90.
\newblock \href {https://doi.org/10.1007/978-3-030-22496-7_5}
  {\path{doi:10.1007/978-3-030-22496-7_5}}.

\bibitem{D2D_Wireless}
H.~Xing, O.~Simeone, S.~Bi, Federated learning over wireless device-to-device
  networks: Algorithms and convergence analysis, arXiv:2101.12704 (Jan 2021).

\bibitem{ram2010d2d}
S.~S. Ram, A.~Nedi{\'c}, V.~V. Veeravalli, Distributed stochastic subgradient
  projection algorithms for convex optimization, Journal of optimization theory
  and applications 147~(3) (2010) 516--545.

\bibitem{Savazzi_Nicoli_Rampa_Kianoush_2020}
S.~Savazzi, M.~Nicoli, V.~Rampa, S.~Kianoush, Federated learning with mutually
  cooperating devices: A consensus approach towards server-less model
  optimization, in: ICASSP 2020 - 2020 IEEE International Conference on
  Acoustics, Speech and Signal Processing (ICASSP), 2020, p. 3937–3941.
\newblock \href {https://doi.org/10.1109/ICASSP40776.2020.9054055}
  {\path{doi:10.1109/ICASSP40776.2020.9054055}}.

\bibitem{Savazzi_Nicoli_Rampa_2020}
S.~Savazzi, M.~Nicoli, V.~Rampa, Federated learning with cooperating devices: A
  consensus approach for massive iot networks, IEEE Internet of Things Journal
  7~(5) (2020) 4641–4654.
\newblock \href {https://doi.org/10.1109/JIOT.2020.2964162}
  {\path{doi:10.1109/JIOT.2020.2964162}}.

\bibitem{Li_Li_Varshney_2022}
C.~Li, G.~Li, P.~K. Varshney, Decentralized federated learning via mutual
  knowledge transfer, IEEE Internet of Things Journal 9~(2) (2022) 1136–1147.
\newblock \href {https://doi.org/10.1109/JIOT.2021.3078543}
  {\path{doi:10.1109/JIOT.2021.3078543}}.

\bibitem{Wang_Wang_Chen_Ji_2021}
J.~Wang, S.~Wang, R.-R. Chen, M.~Ji, Local averaging helps: Hierarchical
  federated learning and convergence analysis, arXiv:2010.12998 (Mar 2021).

\bibitem{Abad_Ozfatura_GUndUz_Ercetin_2020}
M.~S.~H. Abad, E.~Ozfatura, D.~GUndUz, O.~Ercetin, Hierarchical federated
  learning across heterogeneous cellular networks, in: ICASSP 2020 - 2020 IEEE
  International Conference on Acoustics, Speech and Signal Processing (ICASSP),
  2020, p. 8866–8870.
\newblock \href {https://doi.org/10.1109/ICASSP40776.2020.9054634}
  {\path{doi:10.1109/ICASSP40776.2020.9054634}}.

\bibitem{liu2020client}
L.~Liu, J.~Zhang, S.~Song, K.~B. Letaief, Client-edge-cloud hierarchical
  federated learning, in: ICC 2020-2020 IEEE International Conference on
  Communications (ICC), IEEE, 2020, pp. 1--6.

\bibitem{Lin_Hosseinalipour_Azam_Brinton_Michelusi_2021}
F.~P.-C. Lin, S.~Hosseinalipour, S.~S. Azam, C.~G. Brinton, N.~Michelusi,
  Semi-decentralized federated learning with cooperative d2d local model
  aggregations, IEEE Journal on Selected Areas in Communications (2021)
  1–1\href {https://doi.org/10.1109/JSAC.2021.3118344}
  {\path{doi:10.1109/JSAC.2021.3118344}}.

\bibitem{hashemi2021benefits}
A.~Hashemi, A.~Acharya, R.~Das, H.~Vikalo, S.~Sanghavi, I.~Dhillon, On the
  benefits of multiple gossip steps in communication-constrained decentralized
  federated learning, IEEE Transactions on Parallel and Distributed Systems
  33~(11) (2021) 2727--2739.

\bibitem{wang2021edge}
T.~Wang, Y.~Liu, X.~Zheng, H.-N. Dai, W.~Jia, M.~Xie, Edge-based communication
  optimization for distributed federated learning, IEEE Transactions on Network
  Science and Engineering (2021).

\bibitem{ng2021reputation}
J.~S. Ng, W.~Y.~B. Lim, Z.~Xiong, X.~Cao, J.~Jin, D.~Niyato, C.~Leung, C.~Miao,
  Reputation-aware hedonic coalition formation for efficient serverless
  hierarchical federated learning, IEEE Transactions on Parallel and
  Distributed Systems 33~(11) (2021) 2675--2686.

\bibitem{singh2019detailed}
A.~Singh, P.~Vepakomma, O.~Gupta, R.~Raskar, Detailed comparison of
  communication efficiency of split learning and federated learning, arXiv
  preprint arXiv:1909.09145 (2019).

\bibitem{nilsson2018performance}
A.~Nilsson, S.~Smith, G.~Ulm, E.~Gustavsson, M.~Jirstrand, A performance
  evaluation of federated learning algorithms, in: Proceedings of the second
  workshop on distributed infrastructures for deep learning, 2018, pp. 1--8.

\bibitem{hegedHus2021decentralized}
I.~Hegedüs, G.~Danner, M.~Jelasity, Decentralized learning works: An empirical
  comparison of gossip learning and federated learning, Journal of Parallel and
  Distributed Computing 148 (2021) 109--124.

\bibitem{beutel2020flower}
D.~J. Beutel, T.~Topal, A.~Mathur, X.~Qiu, T.~Parcollet, P.~P. de~Gusm{\~a}o,
  N.~D. Lane, Flower: A friendly federated learning research framework,
  arXiv:2007.14390 (2020).

\bibitem{tff}
{T}ensor{F}low {F}ederated --- tensorflow.org,
  \url{https://www.tensorflow.org/federated}.

\bibitem{lai2021fedscale}
F.~Lai, Y.~Dai, X.~Zhu, H.~V. Madhyastha, M.~Chowdhury, Fedscale: Benchmarking
  model and system performance of federated learning, in: Proceedings of the
  First Workshop on Systems Challenges in Reliable and Secure Federated
  Learning, 2021, pp. 1--3.

\bibitem{caldas2018leaf}
S.~Caldas, S.~M.~K. Duddu, P.~Wu, T.~Li, J.~Kone{\v{c}}n{\`y}, H.~B. McMahan,
  V.~Smith, A.~Talwalkar, Leaf: A benchmark for federated settings,
  arXiv:1812.01097 (2018).

\bibitem{li2021federated}
Q.~Li, Y.~Diao, Q.~Chen, B.~He, Federated learning on non-iid data silos: An
  experimental study, arXiv preprint arXiv:2102.02079 (2021).

\bibitem{he2020fedml}
C.~He, S.~Li, J.~So, X.~Zeng, M.~Zhang, H.~Wang, X.~Wang, P.~Vepakomma,
  A.~Singh, H.~Qiu, et~al., Fedml: A research library and benchmark for
  federated machine learning, arXiv:2007.13518 (2020).

\bibitem{konevcny2016federated}
J.~Kone{\v{c}}n{\`y}, H.~B. McMahan, F.~X. Yu, P.~Richt{\'a}rik, A.~T. Suresh,
  D.~Bacon, Federated learning: Strategies for improving communication
  efficiency, arXiv:1610.05492 (2016).

\bibitem{Wang_Han_Leung_Niyato_Yan_Chen_2020}
X.~Wang, Y.~Han, V.~C.~M. Leung, D.~Niyato, X.~Yan, X.~Chen, Convergence of
  edge computing and deep learning: A comprehensive survey, IEEE Communications
  Surveys Tutorials 22~(2) (2020) 869–904.
\newblock \href {https://doi.org/10.1109/COMST.2020.2970550}
  {\path{doi:10.1109/COMST.2020.2970550}}.

\bibitem{tehrani2014device}
M.~N. Tehrani, M.~Uysal, H.~Yanikomeroglu, Device-to-device communication in 5g
  cellular networks: challenges, solutions, and future directions, IEEE
  Communications Magazine 52~(5) (2014) 86--92.

\bibitem{daily2018gossipgrad}
J.~Daily, A.~Vishnu, C.~Siegel, T.~Warfel, V.~Amatya, Gossipgrad: Scalable deep
  learning using gossip communication based asynchronous gradient descent,
  arXiv:1803.05880 (2018).

\bibitem{deng2012mnist}
L.~Deng, The mnist database of handwritten digit images for machine learning
  research, IEEE Signal Processing Magazine 29~(6) (2012) 141--142.

\bibitem{xiao2017fashion}
H.~Xiao, K.~Rasul, R.~Vollgraf, Fashion-mnist: a novel image dataset for
  benchmarking machine learning algorithms, arXiv:1708.07747 (2017).

\bibitem{li2021model}
Q.~Li, B.~He, D.~Song, Model-contrastive federated learning, in: Proceedings of
  the IEEE/CVF Conference on Computer Vision and Pattern Recognition, 2021, pp.
  10713--10722.

\bibitem{guha2019one}
N.~Guha, A.~Talwalkar, V.~Smith, One-shot federated learning, arXiv:1902.11175
  (2019).

\bibitem{7343681}
M.~Condoluci, L.~Militano, A.~Orsino, J.~Alonso-Zarate, G.~Araniti, Lte-direct
  vs. wifi-direct for machine-type communications over lte-a systems, in: 2015
  IEEE 26th Annual International Symposium on Personal, Indoor, and Mobile
  Radio Communications (PIMRC), 2015, pp. 2298--2302.
\newblock \href {https://doi.org/10.1109/PIMRC.2015.7343681}
  {\path{doi:10.1109/PIMRC.2015.7343681}}.

\end{thebibliography}
\bibliographystyle{elsarticle-num}

\end{document}